\pdfoutput=1

\documentclass{article}
\usepackage[utf8]{inputenc}

\usepackage{amsmath,amsfonts}
\usepackage{amssymb}
\usepackage{geometry}
\usepackage{float}
\usepackage{comment}
\usepackage[normalem]{ulem}
\usepackage{multirow}
\useunder{\uline}{\ul}{}
\usepackage[toc,page]{appendix}
\usepackage{cite}
\usepackage[pdftex]{graphicx}
\usepackage{algorithmic}
\usepackage{array}
\usepackage{fixltx2e}
\usepackage{caption}
\usepackage{subcaption}
\usepackage{hyperref}
\usepackage{lscape}

\newcommand\blfootnote[1]{%
  \begingroup
  \renewcommand\thefootnote{}\footnote{#1}%
  \addtocounter{footnote}{-1}%
  \endgroup
}

\newcommand{\ddtheta}{\frac{d}{d\theta}}
\newcommand{\partialdtheta}{\frac{\partial}{\partial\theta}}
\newcommand{\thetai}{{(\theta,i)}}
\newcommand{\thetaj}{{(\theta,j)}}
\newcommand{\Normal}{\mathcal{N}}
\newcommand{\logN}{\log\mathcal{N}}
\newcommand{\fq}{Q}

\newcommand{\nx}{{n_X}}

\newcommand{\ntheta}{{n_\Theta}}

\newcommand{\xp}{{x^\prime}}
\newcommand{\ha}{{\tilde{h}}}
\newcommand{\Ha}{{\tilde{H}}}
\newcommand{\Ra}{{\tilde{R}}}
\newcommand{\resample}{\kappa}

\usepackage[ruled,vlined,linesnumbered]{algorithm2e}
\newcommand{\resampleu}{\nu} 

\title{Efficient Learning of the Parameters of Non-Linear Models using Differentiable Resampling in Particle Filters \footnote{This work has been submitted to the IEEE for possible publication. Copyright may be transferred without notice, after which this version may no longer be accessible.}}

\author{
  Conor Rosato \\
  \and
  Vincent~Beraud \\
  \and 
  Paul Horridge \\
  \and
  Thomas B. Sch\"{o}n \\
  \and
  Simon Maskell \\
}

\date{%
    November 2021
    \blfootnote{C. Rosato, V. Beraud, P. Horridge and S. Maskell are with the Department of Electrical Engineering and Electronics, University of Liverpool, United Kingdom,
    e-mail: (c.m.rosato@liverpool.ac.uk, vincent.beraud@liverpool.ac.uk, p.horridge@liverpool.ac.uk, smaskell@liverpool.ac.uk). Thomas~B.~Sch\"{o}n is with Uppsala University, Sweden, e-mail: (thomas.schon@it.uu.se)}
}

\begin{document}

\maketitle

\begin{abstract}

It has been widely documented that the sampling and resampling steps in particle filters cannot be differentiated. The {\itshape reparameterisation trick} was introduced to allow the sampling step to be reformulated into a differentiable function. We extend the {\itshape reparameterisation trick} to include the stochastic input to resampling therefore limiting the discontinuities in the gradient calculation after this step. Knowing the gradients of the prior and likelihood allows us to run particle Markov Chain Monte Carlo (p-MCMC) and use the No-U-Turn Sampler (NUTS) as the proposal when estimating parameters.

We compare the Metropolis-adjusted Langevin algorithm (MALA), Hamiltonian Monte Carlo with different number of steps and NUTS. We consider three state-space models and show that NUTS improves the mixing of the Markov chain and can produce more accurate results in less computational time.
\end{abstract}

\section{Introduction}\label{sec:intro}

State-Space Models (SSMs) have been used to model dynamical systems in a wide range of research fields (see \cite{doucet2001sequential} for numerous examples). SSMs are represented by two stochastic processes: $\{X_t\}_{t\geq0}$
and $\{Y_t\}_{t\geq0}$ where ${X_t}$ indicates the hidden state which evolves according to a Markov process $p\left(x_t|x_{t-1},  \theta\right)$ and ${Y_t}$ is the observation (both at time $t\geq0$), such that
\begin{equation}
X_t|X_{t-1} \sim\ p\left(x_t|x_{t-1},  \theta\right),
\label{eq:state_estimation}
\end{equation}
\begin{equation}
Y_t|X_t \sim\ p\left(y_t|x_t, \theta\right).
\label{eq:obs_estimation}
\end{equation} 
The initial latent state $X_0$ has initial density denoted $\mu_\theta(x_0)$. The SSM is parameterised by an unknown static parameter $\theta$ contained in the parameter space $\Theta$. The transition and observation densities are given by  (\ref{eq:state_estimation}) and (\ref{eq:obs_estimation}),  respectively. In this paper we focus on Bayesian parameter estimation in SSMs using Particle Markov Chain Monte Carlo (p-MCMC), as first proposed in \cite{andrieu_doucet_holenstein_2010}. This approach combines two Monte Carlo methods that use repeated sampling techniques to obtain numerical estimates of a target distribution $\pi(\theta)$, for which exact inference is intractable. The two methods are Markov Chain Monte Carlo (MCMC), as described in \cite{robert_2015, ravenzwaaij_cassey_brown_2016, Roberts_2004}, and Sequential Monte Carlo (SMC) i.e. a particle filter, as described in \cite{gordon1993novel, arulampalam_maskell_gordon_clapp_2002, 15yearslater}.

MCMC algorithms such as Metropolis-Hastings (M-H) often use random walk sampling within the proposal. Such proposals can struggle to enable the MCMC to reach the stationary distribution when estimating large numbers of parameters. A related issue can occur with Gibbs samplers when the correlation between parameters is high. These issues can result in the sampler getting stuck in local maxima within~$\pi(\theta)$. Hamiltonian Monte Carlo (HMC), as described in \cite{neal2012mcmc}, is an approach that simulates from a problem-specific Hamiltonian system to generate the samples used in the MCMC. HMC has been seen to be effective when estimating parameters in models when the target distribution is complex or multi-modal but is sensitive to hyperparameters which have to be determined by the user. An adaptive version of HMC called the No-U-Turn Sampler (NUTS) \cite{NUTS} automates the selection of these hyperparameters. Probabilistic programming languages such as Stan \cite{stan} and PyMC3 \cite{pymc3} are tools that have been developed to allow users to define and make inferences about probabilistic models using NUTS. 

Particle filters have been used in many areas of research, such as finance \cite{filteringfinance}, disease modelling \cite{filteringdisease} and multiple target tracking \cite{filteringmultipletargets} to infer  time-dependent hidden states. The original contribution of \cite{andrieu_doucet_holenstein_2010} uses a particle filter to calculate an unbiased estimate of the often intractable likelihood for $\theta$. A M-H algorithm with a random walk proposal was used to sample from $\pi(\theta)$. Using such a proposal in p-MCMC will inherit the same issues as described above in the context of MCMC generically. To make use of more sophisticated proposals the gradient of the log-posterior of the parameter, $\theta$, needs to be estimated.

Extensions of the original p-MCMC algorithm have focused on including gradient information when proposing new parameters. Reference \cite{poyiadjis2011particle} shows how to estimate the score (gradient) of the log-likelihood and the observed information matrices at $\theta$ in SSMs using particle filter methods. The two methods proposed run with computational complexity $\mathcal{O}(N)$ and $\mathcal{O}(N^2)$, respectively. The first has a linear computational cost but the performance decreases over time. The second has a computational cost that increases quadratically with the number of particles $N$ but performance does not deteriorate over time, with \cite{del2015uniform} theoretically substantiating this claim. \cite{nemeth2013particle} built on this work to compute these terms with computational complexity $\mathcal{O}(N)$ and avoids the quadratically increasing variance caused by particle degeneracy. In \cite{particleLangevin, inproceedings, Second-order, Dahlin_2014} the authors utilise the previous work of \cite{nemeth2013particle} to recursively estimate the score (gradient) of the log-likelihood at $\theta$. References \cite{particleLangevin} and \cite{inproceedings} include Langevin Monte Carlo (LMC) methods seen in \cite{girolami_calderhead_2011} whilst \cite{Second-order} and \cite{Dahlin_2014} include first- and second-order Hessian information about the posterior in the proposal. Use of the Hessian is shown to improve the mixing of the Markov chain at the stationary phase and decrease the length of burn-in. However, calculating a $d\times d$ matrix of the second-order partial derivatives can become infeasible when the dimensionality, $d$, becomes large. While \cite{Second-order}, \cite{Dahlin_2014} do mention using HMC dynamics within p-MCMC,  to the best of the authors' knowledge, no implementation of this approach  has been described in the literature up to now. 

We aim to complement the recent literature seen in machine learning that addresses the problem of differentiating resampling (see section \ref{sec:diffPFs})). In order to obtain the gradient of the log-likelihood w.r.t $\theta$, the particle filter needs to differentiated. However, it has been noted in \cite{diffPF_1, diffPF_2, diffPF_3} that the stochastic nature of both the sampling and resampling steps, that are inherently part of the particle filter, are not differentiable. As will be explained in more detail later in this paper (in Sections \ref{sect:likelihoodandgradients} and \ref{subsect:particlegradient}), the {\itshape reparameterisation trick} was proposed in \cite{reparam} to reformulate the sampling operation into a differentiable function by sampling a noise vector in advance and defining the likelihood for $\theta$ as being a deterministic function of this sampled noise vector. However, resampling remains problematic since after resampling all weights are equal. More specifically, the gradients cannot be calculated since the new particles' states are not differentiable w.r.t. the weights that are input to resampling. Recent work in machine learning has focused on how to modify the resampling step to make it differentiable.

Using the {\itshape reparameterisation trick} for resampling has been described in \cite{lee2008towards} in the context of fixing the random number seed in every simulation to produce common random numbers (CRN). Our core contribution is to fix the random numbers used within the resampling step so we can condition the input to resampling which results in the subsequent particle derivative calculations being a function of the parent particle. The gradients can then be efficiently estimated and utilised within the framework of p-MCMC and specifically used to calculate gradient-based proposals for $\theta$: more specifically, this allows us to use NUTS as the proposal. Another novel contribution, relative to the previous work on differentiable particle filters in the neural network community, is that we provide full Bayesian parameter estimates (including variances). This differs from the present literature on differentiable particle filtering which focuses exclusively on point-estimates of parameters.  
We also compare NUTS' performance with that achieved by HMC and Metropolis-adjusted Langevin algorithm (MALA).

An outline of the paper is as follows: in Section \ref{sec:modelformulation} we describe a generic particle filter followed by a description of the difficulties associated with differentiating the sampling and resampling steps. We describe how to calculate the likelihood and gradients in Section \ref{sect:likelihoodandgradients}, the methods to propagate the derivatives of the particle weights in Section \ref{sect:calcder} and how we extend the {\itshape reparameterisation trick} to include the stochastic input to resampling in section \ref{sect:resampling}. We test the likelihood and gradient estimates, explain particle-HMC (p-HMC) and particle-NUTS (p-NUTS) and detail comparative numerical results in the context of three applications in Section \ref{sec:testingestimates}. Concluding remarks are described in Section \ref{sec:conclusion}.

\section{Particle filtering background}
\label{sec:modelformulation}
Assume we have considered $t$ time-steps, obtaining data at each increment of $t$ given by $y_{1:t}$. The state sequence $x_{1:t}$ grows with time where $x_t$ has $n_x$ dimensions. The dynamics and likelihood are parameterised by $\theta$ (which has $n_\theta$ dimensions) such that

\begin{align}
p\left(y_{1:t},x_{1:t}|\theta\right) = p(y_1&|x_1,\theta)p\left(x_1|\theta\right) \\ \nonumber
&\times \prod_{\tau=2}^{t}p\left(y_\tau|x_\tau,\theta\right)p\left(x_\tau|x_{\tau-1},\theta\right).
\end{align}
If $\theta$ is known, we can run a (conventional) particle filter.

\subsection{Particle Filter}
\label{ParticleFilter}

At every timestep $t$, the particle filter draws $N$ samples (particles) from a proposal distribution, $q\left(x_{1:t}|y_{1:t},\theta\right)$, which is parameterised by the sequence of states and measurements. The samples are seen as statistically independent and each represents a different hypothesis for the sequence of states of the system. The $i$th sample has an associated weight, $w_{t}^{(\theta,i)}$, which indicates the relative importance of each of the $N$ samples. The weights at $t=0$ are set to be $1/N$. The proposal distribution is constructed recursively as
\begin{equation}
q\left(x_{1:t}|y_{1:t},\theta\right) = q\left(x_1|y_1,\theta\right)\prod_{\tau=2}^t q\left(x_\tau|x_{\tau-1},y_\tau,\theta\right),
\end{equation}
such that we can pose an estimate with respect to the joint distribution, $p\left(y_{1:t},x_{1:t}|\theta\right)$, as follows:
\begin{equation}
\int p\left(y_{1:t},x_{1:t}|\theta\right)f\left(x_{1:t}\right) dx_{1:t} \approx \frac{1}{N}\sum_{i=1}^N w_{1:t}^{(\theta,i)}f\left(x_{1:t}^{(i)}\right).\label{eq:jointexp}
\end{equation}

This is an unbiased estimate, where (for $t>$1)
\begin{align}
w_{1:t}^{(\theta,i)}=&\frac{p\left(y_1|x_1^{(\theta,i)},\theta\right)p\left(x_1^{(\theta,i)}|\theta\right)}{q\left(x^{(\theta,i)}_1|y_1,\theta\right)}\nonumber\\
&\times \frac{\prod_{\tau=2}^{t}p\left(y_\tau|x_\tau^{(\theta,i)},\theta\right)p\left(x_\tau^{(\theta,i)}|x^{(\theta,i)}_{\tau-1},\theta\right)}{\prod_{\tau=2}^t q\left(x_\tau^{(\theta,i)}|x_{\tau-1}^{(\theta,i)},y_\tau,\theta\right)} \\ 
=& w_{1:t-1}^{(\theta,i)}\frac{p\left(y_t|x_t^{(\theta,i)},\theta\right)p\left(x_t^{(\theta,i)}|x^{(\theta,i)}_{t-1},\theta\right)}{q\left(x_t^{(\theta,i)}|x_{t-1}^{(\theta,i)},y_t\right)},
\end{align}
and is a recursive formulation for the unnormalised weight, $w_{1:t}^{(\theta,i)}$, with incremental weight
\begin{equation}
\sigma\left(x_t^\thetai, x_{t-1}^\thetai, \theta\right)=\frac{p\left(y_t|x_t^{(\theta,i)},\theta\right)p\left(x_t^{(\theta,i)}|x^{(\theta,i)}_{t-1},\theta\right)}{q\left(x_t^{(\theta,i)}|x_{t-1}^{(\theta,i)},y_t\right)}.
\label{eq:incrementalweight}
\end{equation}
For $t$=1

\begin{equation}
\sigma\left(x_{1:1}^{(\theta,i)}\right) = \frac{p\left(y_1|x_1^{(\theta,i)},\theta\right)p\left(x_1^{(\theta,i)}|\theta\right)}{q\left(x_1^{(\theta,i)}|y_1\right)}.
\end{equation}

\subsection{Choice of proposal}
\label{whichproposal}

Three commonly used options for the proposal distribution are:

\begin{enumerate}
	\item Using the dynamic model as the proposal
	\begin{align}
	q\left(x_t^{(\theta,i)}|x_{t-1}^{(\theta,i)},y_t\right)=p\left(x_t^{(i)}|x^{(\theta,i)}_{t-1},\theta\right),
	\label{eq:priorproposal}
	\end{align}
	which simplifies the weight update to
	\begin{align}
	w_{1:t}^{(\theta,i)}=p\left(y_t|x_t^{(\theta,i)},\theta\right)w_{1:t-1}^{(\theta,i)}.
	\label{eq:priorproposalweight}
	\end{align}
	\item In certain situations it is possible to use the ``optimal'' proposal which is
	\begin{align}
	q\left(x_t^{(\theta,i)}|x_{t-1}^{(\theta,i)},y_t\right) = p\left(x_t^{(\theta,i)}|x_{t-1}^{(\theta,i)},y_t\right)
	\label{eq:optimalproposalpropogate}
	\end{align}
	with weights updated according to 
	\begin{align}
	w_{1:t}^{(\theta,i)}=p\left(y_t|x_{t-1}^{(\theta,i)},\theta\right)w_{1:t-1}^{(\theta,i)}.
	\label{eq:optimalproposalweight}
	\end{align}
 Note that the term ``optimal'' in this context of the particle proposal means that the variance of the incremental particle weights at the current time-step is minimized. In fact, this variance is zero since the weight in~(\ref{eq:optimalproposalweight}) is independent of $x_{t}$ (as explained in \cite{1336055731117}).
	\item Using the Unscented Transform, as explained in \cite{kalmanFil}.
\end{enumerate} 

\subsection{Estimation with Respect to the Posterior}
It is often the case that we wish to calculate estimates with respect to the posterior, $p\left(x_{1:t}|y_{1:t},\theta\right)$, which we can calculate as follows:
\begin{align}
\int p\left(x_{1:t}|y_{1:t},\theta\right)&f\left(x_{1:t}\right) dx_{1:t} \\
= &\int \frac{p\left(y_{1:t},x_{1:t}|\theta\right)}{p\left(y_{1:t}|\theta\right)}f\left(x_{1:t}\right) dx_{1:t},
\end{align}
\normalsize

\begin{equation}
p\left(y_{1:t}|\theta\right) = \int  p\left(y_{1:t},x_{1:t}|\theta\right)dx_{1:t}\approx  \frac{1}{N}\sum_{i=1}^N w_{1:t}^{(\theta,i)}\label{eq:likelihood}
\end{equation}

in line with (\ref{eq:jointexp}), such that
\begin{align}
\int  p&\left(x_{1:t}|y_{1:t},\theta\right)f\left(x_{1:t}\right)dx_{1:t} \nonumber \\
& \approx \frac{1}{\frac{1}{N}\sum_{i=1}^N {w}_{1:t}^{(\theta,i)}}
\frac{1}{N}\sum_{i=1}^N {w}_{1:t}^{(\theta,i)}f\left(x_{1:t}^{(\theta,i)}\right)  \\ 
&= \sum_{i=1}^N \tilde{w}_{1:t}^{(\theta,i)}f\left(x_{1:t}^{(\theta,i)}\right),
\label{eq:a}
\end{align}
where
\begin{equation}
\tilde{w}_{1:t}^{(\theta,i)} = \frac{w_{1:t}^{(\theta,i)}}{\sum_{j=1}^N w_{1:t}^{(\theta,j)}}
\label{eq:normalisedweights}
\end{equation}
are the normalised weights.

Equation \eqref{eq:a} is a biased estimate since it is a ratio of estimates, in contrast with (\ref{eq:jointexp}).

\subsection{Resampling}
\label{Resampling}
The algorithm described up to now is called the sequential importance sampling (SIS) algorithm. As time evolves, the normalised weights will become increasingly skewed such that one of the weights given by \eqref{eq:normalisedweights} becomes close to unity and the others approach zero. This is an inevitability and cannot be avoided \cite{doucet2000sequential}. As well as the number of effective samples, $N_{\text{eff}}$, eventually becoming 1, most of the computational effort will be expended on particles that have very little contribution to the overall estimate.

It is often suggested that monitoring $N_{\text{eff}}$, can be used to identify the need to resample, where
\begin{equation}
N_{\text{eff}} = \frac{1}{\sum_{i=1}^N \left(\tilde{w}_{1:t}^{(\theta,i)}\right)^2}.
\label{eq:neff}
\end{equation}
There are many resampling methods, some of which are outlined and evaluated in \cite{resamplingmethods} but they all share the same purpose---stochastically replicate particles with higher weights whilst eliminating ones with lower weights. Multinomial resampling is commonly used and involves drawing from the current particle set $N$ times proportional to its weight. The associated distribution is defined by
\begin{equation}
\tilde{w}_{1:t}^{(\theta,i)} \ \ \text{for} \ \  i=1,\ldots ,N.
\label{eq:resampleparticles}
\end{equation}
To keep the total unnormalised weight constant (such that the approximation \eqref{eq:likelihood} is the same immediately before and after resampling), we assign each newly-resampled sample an unnormalised weight
\begin{equation}
\frac{1}{N}\sum_{i=1}^N w_{1:t}^{(\theta,i)}.
\label{eq:normalisedafterresampling}
\end{equation}
Note this is such that the normalised weights after resampling are $\frac{1}{N}$.

\section{Calculating the likelihood and gradients}\label{sect:likelihoodandgradients}

We pose the calculation of the likelihood of the parameter as the calculation of the approximation in (\ref{eq:likelihood}), (ie the sum of unnormalised particle filter weights), with $t=T$.  

Differentiating the weights gives an approximation to the gradient of the likelihood\footnote{We note that this approach differs from that advocated in~\cite{inproceedings,Second-order,Dahlin_2014}, which use a Fixed-Lag filter (with a user-specified lag) to approximate the derivatives. In contrast, we explicitly calculate the derivatives of the approximation to the likelihood.}:
\begin{eqnarray}
	\ddtheta p(y_{1:t} | \theta) & \approx & \frac{1}{N} \sum_{i=1}^N \ddtheta w_{1:t}^{(\theta,i)}. \label{eqn:dlikweightsum}
\end{eqnarray}
For numerical stability, it is typically preferable to propagate values in logs. Applying the Chain Rule to (\ref{eq:likelihood}) and (\ref{eqn:dlikweightsum}) gives

\begin{align}
	\ddtheta \log p(y_{1:t} | \theta) &\approx \frac{1}{N} \frac{1}{p(y_{1:t} | \theta)}\sum_{i=1}^N w_{1:t}^{(\theta,i)}\ddtheta \log w_{1:t}^{(\theta,i)}
		\\ &\approx \frac{1}{N}
	\sum_{i=1}^N \tilde{w}_{1:t}^{(\theta,i)}\ddtheta \log w_{1:t}^{(\theta,i)}.	\label{eqn:dloglik}
\end{align}

Note that in (\ref{eqn:dloglik}), the weights outside the logs are normalised, while the weights inside are not. 
The log weights can be calculated recursively as
\begin{align}	\label{eqn:weightupdate}
	\log w_{1:t}^\thetai &= \log w_{1:t-1}^\thetai + \log \sigma\left(x_t^\thetai, x_{t-1}^\thetai, \theta\right),
\end{align}
so
\begin{align}	\label{eqn:logweightderivative}
	\ddtheta \log w_{1:t}^\thetai &= \ddtheta  \log w_{1:t-1}^\thetai + \ddtheta \log \sigma\left(x_t^\thetai, x_{t-1}^\thetai, \theta\right),
\end{align}
where
\begin{align}
	\ddtheta &\log\sigma\left(x_t^\thetai, x_{t-1}^\thetai, \theta\right) \nonumber\\ 
	=& \ddtheta \log p\left(x_t^\thetai | x_{t-1}^\thetai, \theta\right)
		+ \ddtheta \log p\left(y_t | x_{t-1}^\thetai\right) \nonumber\\  
		&- \ddtheta \log q\left(x_t^\thetai | x_{t-1}^\thetai, \theta, y_t\right).\label{eq:derivativeofproposal}
\end{align}

So, if we can differentiate the single measurement likelihood, transition model and proposal, we can calculate (an approximation to) the derivative of the log-likelihood for the next time step, thus recursively approximating the log-likelihood derivatives for each time step. While this is true, there are some challenges involved, which we now discuss in turn.

If the particle filter is using the transition model as the dynamics (as in (\ref{eq:priorproposal})), the likelihood in the weight update does not explicitly depend on $\theta$ and we might initially suppose that $d\log \sigma/ d\theta = 0$. If this were the case, an induction argument using (\ref{eqn:logweightderivative}) would show that the weight derivatives were always zero and therefore give an approximation of zero for the gradient of the likelihood for $\theta$. This seems intuitively incorrect. Indeed, the flaw in this reasoning is that, in fact, the likelihood (somewhat implicitly) does depend on $\theta$ since $x_{t}^\thetai$ depends on $\theta$. Applying the Chain Rule gives
\begin{align}
	 \ddtheta \log p\left(y_t | x_{t}^\thetai\right) &= \left.\frac{d}{d x}\log p\left(y_t | x \right)\right\rvert_{x = x_{t}^\thetai}
	 	\frac{d}{d\theta}x_{t}^{\thetai}.\label{eq:chainruleforlikelihood}
\end{align}

Since $x_{t}^\thetai$ is a random variable sampled from the proposal, we use the \emph{reparameterisation trick}\cite{reparam}: we consider the derivative for a fixed random number seed. More precisely, let $\epsilon^{(i)}_t$ be the vector of standard $\Normal(0,1)$ random variables used when sampling from the proposal such that, if $\epsilon^{(i)}_{t}$ is known, then $x_t^{(\theta,i)}$ is a deterministic function (that can be differentiated) of $x_{t-1}^{(\theta,i)}$. We then consider
\begin{align}
    \frac{d}{d\theta} p\left(y_{1:t}|\theta\right) =&  \frac{d}{d\theta}\int p\left(y_{1:t},\epsilon_{1:t}|\theta\right)d\epsilon_{1:t}\\
    =&\int \frac{d}{d\theta} p\left(y_{1:t},\epsilon_{1:t}|\theta\right)d\epsilon_{1:t}\\
    \approx & \frac{1}{N} \sum_{i=1}^N \frac{d}{d\theta} p\left(y_{1:t}|\epsilon^{(i)}_{1:t},\theta\right)\label{eq:reparameterisation}
\end{align}
where $\epsilon^{(i)}_{1:t}\sim p(\epsilon_{1:t})$ is considered fixed and, most importantly, (\ref{eq:reparameterisation}) then involves differentials that can be calculated.

As a simple example, consider sampling from the dynamics with a random walk proposal and $\theta$ being the standard deviation of the process noise.  This is such that
\begin{eqnarray}
	x_t^{\thetai} 
	& = & x_{t-1}^\thetai + \theta\epsilon^{(i)}_t
\end{eqnarray}
so
\begin{eqnarray}
	\frac{d}{d\theta} x_t^{\thetai}& = & \frac{d}{d\theta}x_{t-1}^\thetai + \epsilon^{(i)}_t
\end{eqnarray}
which can be calculated recursively and then used to calculate~(\ref{eq:chainruleforlikelihood}). 

More generally, the derivatives of the weight are non-zero and, to calculate these derivatives, we have to propagate the particle derivatives $dx_{t}^\thetai/d\theta$.

\section{Calculating the derivatives} \label{sect:calcder}

In order to propagate the derivatives of the particle weights we need to calculate:
\begin{itemize}
	\item The particle derivatives,
	\begin{eqnarray}
	 	 \frac{dx_{t}^\thetai}{d\theta}.
	\end{eqnarray}
	\item The derivatives of the proposal pdfs,
	\begin{eqnarray}
		\ddtheta \log q\left(x_t^\thetai | x_{t-1}^\thetai, \theta, y_t\right).
	\end{eqnarray}
	\item The derivatives of the prior log pdfs,
	\begin{eqnarray}
	 	 \ddtheta \log p\left(x_t^\thetai | x_{t-1}^\thetai, \theta\right).
	\end{eqnarray}
	\item The derivatives of the single measurement likelihood log pdfs,
	\begin{eqnarray}
	 \ddtheta \log p\left(y_t | x_{t-1}^\thetai\right).
	\end{eqnarray}
\end{itemize}

In this section, we show how to calculate these derivatives in turn.

\subsection{Derivative of New Particle States}	\label{subsect:particlegradient}

We now describe how to calculate $d x_{t-1}^\thetai/d \theta$. Suppose the proposal takes the following form:
\begin{align}
	q\Big(x_t^\thetai | &x_{t-1}^\thetai, \theta, y_t\Big) \\ \nonumber
	&= \Normal\left(x_t^\thetai; \mu\left(x^\thetai_{t-1}, \theta, y_t\right),
		C\left(x^\thetai_{t-1}, \theta, y_t\right)\right)
		\label{eqn:proposal}
\end{align}
where $\mu\left(.\right)$ and $C\left(.\right)$ are functions of the old particle state, the measurement and the parameter. Such a generic description can articulate sampling from the prior,
or defining a proposal using a Kalman filter with the predicted mean and covariance given by the motion model.

If we sample the proposal noise $\epsilon^{(i)}_t \sim \Normal(\cdot; 0, I_\nx)$ in advance, the new particle states can be written as a deterministic function
\begin{align}
	x_t^\thetai &= f(x_{t-1}^\thetai, \theta, y_t, \epsilon^{(i)}_t) \\
	&\triangleq \mu(x^\thetai_{t-1}, \theta, y_t) + \sqrt{C(x^\thetai_{t-1}, \theta, y_t)}\times \epsilon^{(i)}_t. \label{eqn:sampledeterministic}
\end{align}
We would like to compute the derivative of this w.r.t the parameter. Care must be taken however, since $x^\thetai_{t-1}$ is itself a 
function of $\theta$ (if a different $\theta$ was chosen, a different $x^\thetai_{t-1}$ would have been sampled).
\begin{eqnarray}
	\frac{dx_t^\thetai}{d\theta} & = & \frac{d}{d\theta}f(x_{t-1}^\thetai, \theta, y_t, \epsilon^{(i)}_t) \label{eqn:df} \\ & = &
		\frac{\partial f}{\partial x_{t-1}^\thetai}\frac{d x_{t-1}^\thetai}{d \theta} + \frac{\partial f}{\partial\theta} \label{eqn:partialf}
		\frac{d \theta}{d \theta} \\ & = & 
		\frac{\partial f}{\partial x_{t-1}^\thetai}\frac{d x_{t-1}^\thetai}{d \theta} + \frac{\partial f}{\partial\theta}.	\label{eqn:dxk}
\end{eqnarray}
Note that $df/d\theta$ in (\ref{eqn:df}) is not the same as $\partial f/\partial\theta$ in (\ref{eqn:partialf}) --- see Appendix \ref{app:partial} for the distinction.
Note also that the terms here are matrix-valued in general: $\partial f/\partial x_{t-1}^i$ is an $\nx\times\nx$ matrix, and $d x_{t-1}^i/d\theta$
and $\partial f/\partial\theta$ are $\nx\times\ntheta$ matrices.

Also note that for $t\geq 2$, $x_{t-1}^{\thetai}$ implicitly depends on $x_{t-2}^{\thetai}$, which itself depends on $\theta$. Hence we need the total
derivative $d x_{t-1}^\thetai/d \theta$.

\subsection{Derivative of Proposal} \label{subsect:dproposal}

To differentiate the log proposal pdf, we note that we can write it as
\begin{eqnarray}
	\log q\left(x_t^\thetai | x_{t-1}^\thetai, \theta, y_t\right) = \fq\left(x_{t-1}^\thetai, \theta, y_t, \epsilon^{(i)}_t\right)
\end{eqnarray}
where (dropping the fixed values $\epsilon^{(i)}_t$ and $y_t$ for notational convenience)
\begin{align}
	\fq\left(x_{t-1}^\thetai, \theta\right) &\triangleq \log q\left(f(x_{t-1}^\thetai, \theta) | x_{t-1}^\thetai\right) \\ &=
		\logN\left(f(x_{t-1}^\thetai, \theta); \mu(x^\thetai_{t-1}, \theta), C(x^\thetai_{t-1}, \theta)\right)
\end{align}
where we emphasise again that we assume the proposal is Gaussian. We then get
\small
\begin{align}
	 \frac{d}{d\theta}\fq(x_{t-1}^\thetai, \theta) =&
		\frac{\partial}{\partial f}\logN(f; \mu, C)\left(
			\frac{d f}{d\theta} + \frac{d\mu}{d\theta} + \frac{dC}{d\theta}\right)  \\
	 =&\frac{\partial}{\partial f}\logN(f; \mu, C)\left(
			\frac{\partial f}{\partial x^\thetai_{t-1}}\frac{d  x^\thetai_{t-1}}{d \theta} + \frac{\partial f}{\partial\theta}\right) 
			\nonumber \\
	 &+\frac{\partial}{\partial \mu}\logN(f; \mu, C)\left(
			\frac{\partial \mu}{\partial x^\thetai_{t-1}}\frac{d  x^\thetai_{t-1}}{d \theta} + \frac{\partial \mu}{\partial\theta}\right)
			\nonumber \\
	 &+\frac{\partial}{\partial C}\logN(f; \mu, C)\left(
			\frac{\partial C}{\partial x^\thetai_{t-1}}\frac{d x^\thetai_{t-1}}{d \theta} + \frac{\partial C}{\partial\theta}\right).
			\label{eqn:proposalderivative}
\end{align}
\normalsize
where we denote $\mu=\mu(x^\thetai_{t-1}, \theta)$  and $C=C(x^\thetai_{t-1}, \theta)$ for brevity. The derivatives of $\logN(f; \mu, C)$ are given in Appendix \ref{app:normalderiv}.

\subsection{Derivative of the Prior} \label{subsect:dprior}

We now describe how to calculate $\ddtheta \log p\left(x_t^\thetai | x_{t-1}^\thetai, \theta\right)$. Let
\begin{align}
	P(x_{t-1}^\thetai, \theta,  y_t, \epsilon^{(i)}_t) &\triangleq \log p\left(f\left(x_{t-1}^\thetai, \theta, y_t, \epsilon^{(i)}_t\right)  | x^\thetai_{t-1}, \theta
		\right) \\ &=
		\logN\left(f\left(x_{t-1}^\thetai\right); a\left(x_{t-1}^\thetai, \theta\right), \Sigma(\theta)\right)
\end{align}
where here we assume that the transition model has additive Gaussian noise that is independent of $x_{t-1}^\thetai$. Then
\begin{align}
	\ddtheta P(x_{t-1}^\thetai, \theta) =& \frac{\partial}{\partial f}\logN(f; a, \Sigma)\left(
			\frac{\partial f}{\partial x^\thetai_{t-1}}\frac{d x^\thetai_{t-1}}{d\theta} +
				\frac{\partial f}{\partial\theta}\right) \nonumber \\
		&+\frac{\partial}{\partial a}\logN(f; a, \Sigma)\left(
			\frac{\partial a}{\partial x^\thetai_{t-1}}\frac{d x^\thetai_{t-1}}{d\theta} +
			\frac{\partial a}{\partial\theta}\right) \nonumber \\
		&+\frac{\partial}{\partial \Sigma}\logN(f; a, \Sigma)\left(
			\frac{\partial\Sigma}{\partial\theta}\right).	\label{eqn:priorderivative}
\end{align}
where we denote that $a=a\left(x_{t-1}^\thetai, \theta\right)$ and $\Sigma=\Sigma(\theta)$ for brevity. Note that this makes clear that since the realisation of the sampled particles, $x_t^\thetai$, are, in general, dependent on $\theta$, (\ref{eqn:priorderivative}) includes $\frac{d x^\thetai_{t-1}}{d\theta}$. Also note that these derivatives of $\logN(f; a, \Sigma)$ are evaluated at $a(x^\thetai_{t-1})$ (the prior mean) and not at  $\mu$ (the proposal mean) as was the case in (\ref{eqn:proposalderivative}).

\subsection{Derivative of the Likelihood} \label{subsect:dlikelihood}

We now describe how to calculate $\ddtheta \log p\left(y_t | x_{t-1}^\thetai\right)$. Let
\begin{align}
    	L(x^\thetai_t, \theta, y_t) &\triangleq \log p\left(y_t | x^\thetai_t, \theta\right) \\ &= \logN\left(y_t; h(x^\thetai_t, \theta), R(\theta)\right)
\end{align}
where we assume that the likelihood is Gaussian with a variance that is independent of $x^\thetai_t$. Then
\small
\begin{align}
	\frac{d}{d\theta} L\left(x_t^\thetai, \theta, y_t\right) =&
		\frac{\partial}{\partial h}\logN(y_t; h, R)\left(\frac{\partial h}{\partial x^\thetai_t}\frac{d x^\thetai_t}{d \theta} +
			\frac{\partial h}{\partial \theta}\right) \nonumber \\ &+ \frac{\partial}{\partial R}\logN(y_t; h, R)\frac{d R}{d \theta}
\end{align}
\normalsize
where we denote $h=h(x^\thetai_t, \theta)$ and $R=R(\theta)$ for brevity.

\section{Resampling for a Differentiable Particle Filter}	\label{sect:resampling}

Unlike a standard particle filter, we also need to resample the weight derivatives
\begin{eqnarray}
	\ddtheta w^{\thetai}_{1:t}
\end{eqnarray}
as well as the particle derivatives
\begin{eqnarray}	
	\ddtheta x^{\thetai}_t.
\end{eqnarray}
Let
\begin{eqnarray}
	c^{\thetai}_t = \frac{\sum_{j=1}^i w_{1:t}^{(j,\theta)}}{\sum_{j=1}^N w_{1:t}^{(j,\theta)}},
\end{eqnarray}
be the normalised cumulative weights and the index sampled for particle $i$ be given by
\begin{align}
	\resample_i= \resample\left(\resampleu_t^i, w_{1:t}^{1:N}\right) = \sum_{j=0}^{N-1}\left[\resampleu_t^i > c^{(j,\theta)}_t \right]
\end{align}
where $\resampleu_t^i \sim \mbox{Uniform}((0, 1])$ are independent for each particle and time step. Note that the particle indices are sampled according to a Categorical distribution giving a Multinomial resampler, where each index is resampled with probability proportional to its weight. Other resampling schemes are possible and could reduce the variance of any estimates.

The resampled weights are set to be the same as each other, while preserving the original sum:
\begin{eqnarray}
	x^{\prime\thetai}_t & = & x_t^{(\theta, \kappa_i)}, \\
	w^{\prime\thetai}_{1:t} & = & \frac1N \sum_{j=1}^N w_{1:t}^{\thetaj}.	\label{eqn:weightresample}
\end{eqnarray}
From (\ref{eqn:weightresample}), it is clear that
\begin{eqnarray}
	\ddtheta w^{\prime\thetai}_{1:t} & = & \frac1N \sum_{j=1}^N \ddtheta w_{1:t}^{\thetaj}. \label{eqn:weightdiffresample}
\end{eqnarray}
In order to convert this to log weights, applying the Chain Rule gives
\begin{align}
	\ddtheta \log w^{\prime\thetai}_{1:t} &= \frac1N\frac1{w^{\prime\thetai}_{1:t}}\sum_{j=1}^N w_{1:t}^{\thetaj}\ddtheta \log w_{1:t}^{\thetaj}
		\\ &=
	\sum_{j=1}^N \tilde{w}_{1:t}^{\thetaj}\ddtheta \log w_{1:t}^{\thetaj}
\end{align}
where $\tilde{w}_{1:t}^{\thetaj}$ are the normalised weights.

To get the particle gradient note that (where $\resample$ is differentiable),
\begin{align}
	\ddtheta x^{\prime\thetai}_t =&
		\frac{\partial}{\partial\resample}x_t\left(\theta, \resample\left(\resampleu_t^i, w_{1:t}^{1:N}\right)\right) \frac{\partial}{\partial\theta}
			\resample\left(\resampleu_t^i, w_{1:t}^{1:N}\right) \nonumber \\ 
			&+ \frac{d}{d\theta}x_t\left(\theta, \resample\left(\resampleu_t^i, w_{1:t}^{1:N}\right)\right)
\end{align}
Since $\partial\resample/\partial\theta = 0$ except where
\begin{eqnarray}
	\resampleu_t^i & = & c^{(\theta,j)}_t \mbox{ for some $i, j=1,\ldots,N$}
\end{eqnarray}
then
\begin{eqnarray}
	\ddtheta x^{\prime\thetai}_t & = & \frac{d}{d\theta}x_t^{(\theta, \kappa_i)}
\end{eqnarray}
almost surely, so the derivative is obtained by taking the derivative of the parent particle.

\subsection{Discontinuities after a Resampling Realisation}\label{sec:discont}

\begin{figure}[tp]
\centering
\begin{subfigure}[b]{0.5\textwidth}
\includegraphics[width=1\linewidth]{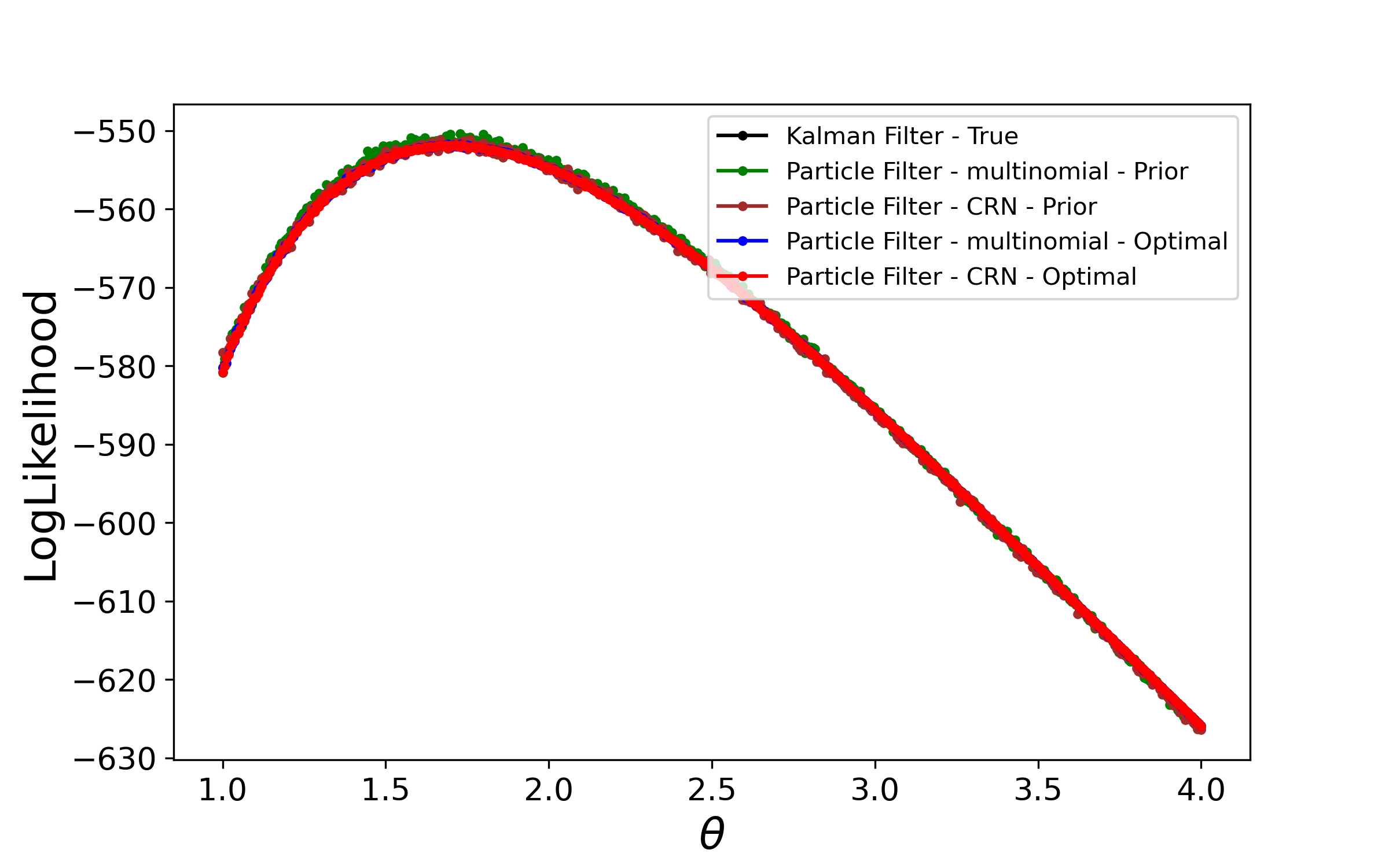}
\caption{}
\end{subfigure}
\hspace{-0.5cm}
\begin{subfigure}[b]{0.5\textwidth}
\includegraphics[width=1\linewidth]{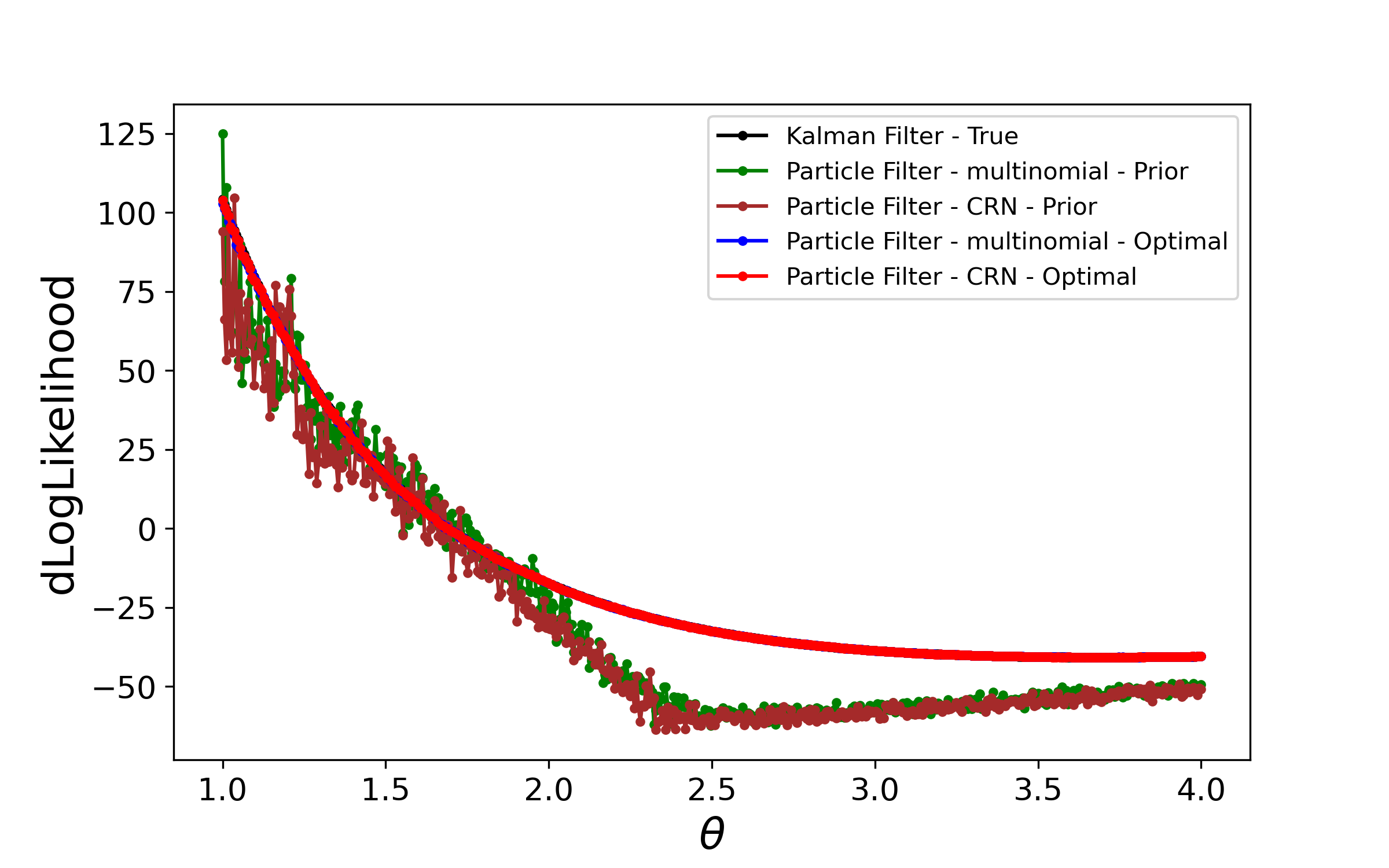}
\caption{}
\end{subfigure}
\hfill
\centering
\begin{subfigure}[b]{0.9\textwidth}
\includegraphics[width=1\linewidth]{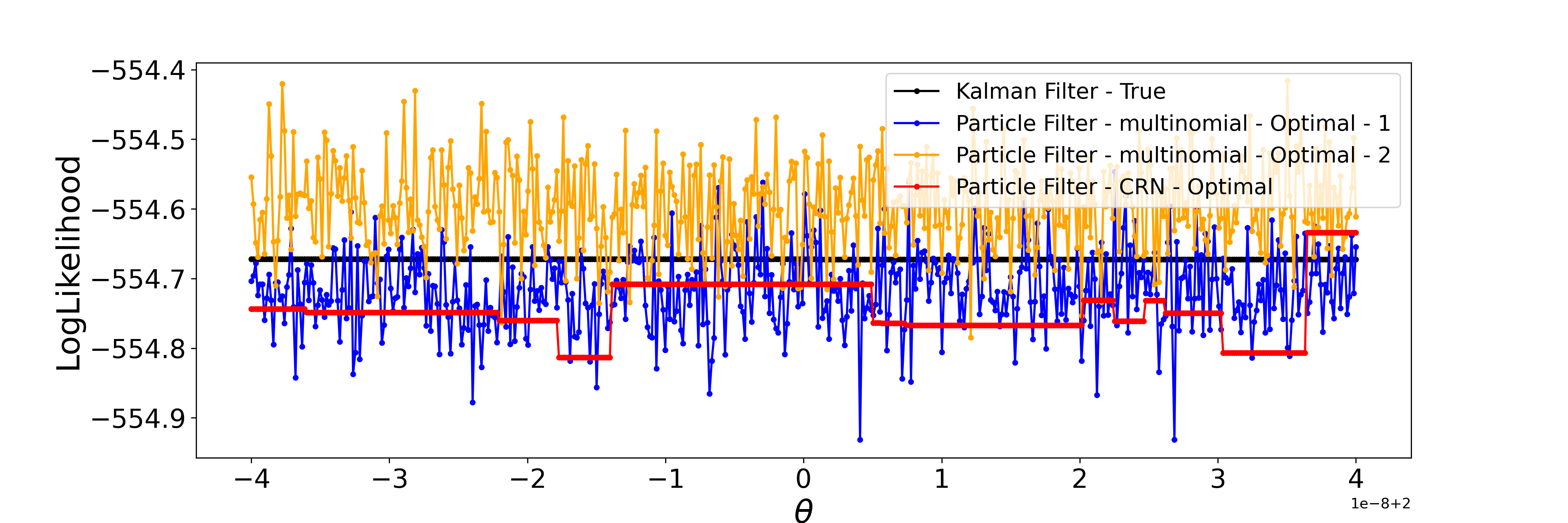}
\caption{}
\end{subfigure}
\hfill
\centering
\begin{subfigure}[b]{0.9\textwidth}
\includegraphics[width=1\linewidth]{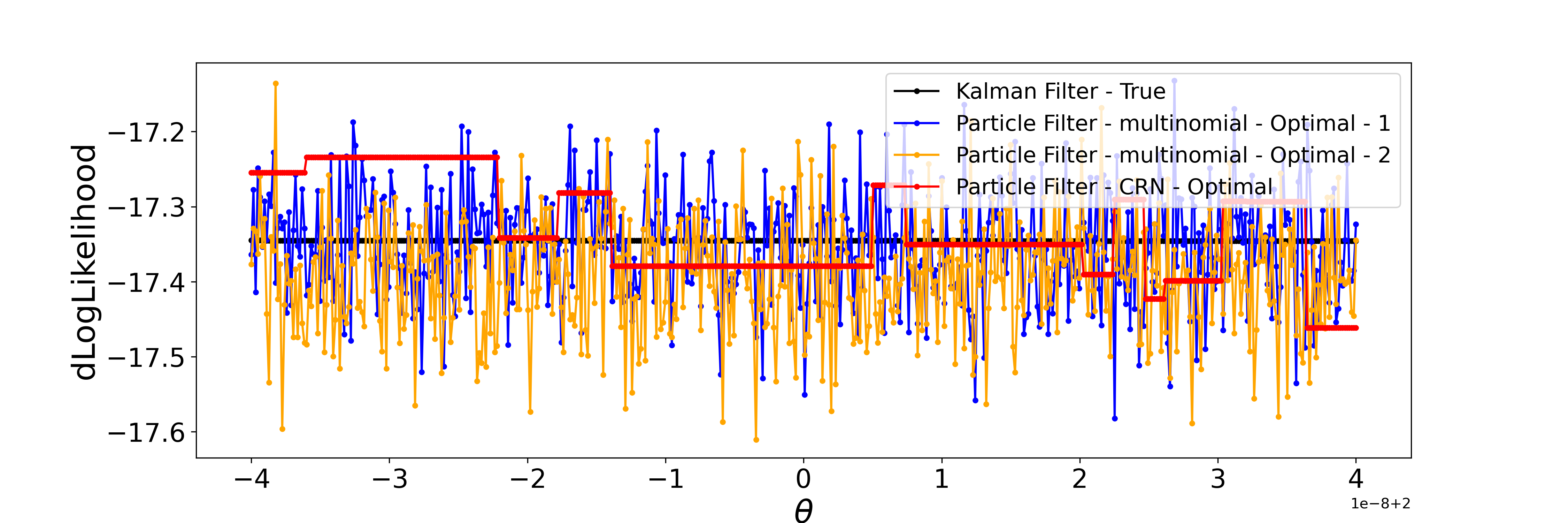}
\caption{}
\end{subfigure}
\caption{Plots of the log-likelihood (a), gradient of the log-likelihood w.r.t. $\theta$ (b), a zoomed in section of the log-likelihood plot (c) and associated gradient (d). All plots: The true values given from the Kalman Filter (black), particle filter using {\itshape reparameterisation trick} resampling (red) and multinomial resampling (blue and orange).}
\label{fig:KF_plots}
\end{figure}

Sampling with CRN results in a ``deterministic" function $f(\theta)$ i.e. evaluating the function twice results in the same output. Sampling without CRN would result in different outputs. 
A discontinuity occurs in the estimate of the log-likelihood in Figure \ref{fig:KF_plots} (a) when two values of $\theta$ cause a different resampling realisation to occur. On one side of the discontinuity, for some range of values of $\theta$, the resampling happens at the same times and all particles have the same parents at all resampling events: the particles for different values of $\theta$ all share a single family tree. When $\theta$ is changed to the other side of the discontinuity the resampling realisations change and so the family tree also changes. Since the approximation to the likelihood (and its gradient) is a function of the family tree, the change in family tree results in a discontinuity in the likelihood approximation.

The best approach to limiting these discontinuities is to come up with a high-performance proposal. This is exemplified when comparing results obtained using the prior with those obtained using the optimal proposal (which can, in some settings, be derived using a Kalman Filter). We choose such a simple example to demonstrate this point. Suppose that the state is a single real number, with a motion model which is a random walk with zero initial mean and the standard deviation of both the initial state and each subsequent propagation is $\theta$. Given the state, the measurements are not dependent on $\theta$ and are equal to the target state plus errors of known and fixed variance, $R$:

\begin{equation}
p\left(x_{1}\right)=\mathcal{N}\left(x_{1} ; 0, \theta^{2}\right),
\end{equation}

\begin{equation}
p\left(x_{k} \mid x_{k-1}, \theta\right)=\mathcal{N}\left(x_{k} ; x_{k-1}, \theta^{2}\right), 
\end{equation}

\begin{equation}
p\left(y_{k} \mid x_{k}, \theta\right)=\mathcal{N}\left(y_{k} ; x_{k}, R\right).
\end{equation}

\noindent We show in Appendix \ref{app:diffkalman} how to calculate the mean and covariance of the optimal proposal for each particle $x_{t-1}^i$, as well as the necessary derivatives.

We run Algorithm \ref{alg:the_alg_1} and compute an estimate of the log-likelihood and associated gradient across a range of 500 values of $\theta$, equally spaced from 1 to 4. The true value is $\theta$=2. We consider $N=2000$ and $T=250$ observations. 

The log-likelihood and gradient of the log-likelihood w.r.t $\theta$, at each instance of $\theta$, can be seen in Figures \ref{fig:KF_plots} (a) and (b), respectively. Figures \ref{fig:KF_plots} (c) and (d) are zoomed in instances of these plots that include results from the Kalman Filter, two runs of multinomial resampling (to indicate the difference in results when running the simulation twice) and CRN resampling. From looking at Figure \ref{fig:KF_plots} (a), there are no obvious differences in the graphs of the log-likelihood for the estimates given by the Kalman Filter and the estimates produced by the particle filter when using different combinations of the prior and optimal proposal with using CRN and multinomial resampling. However, there is a notable difference when comparing the gradient of the log-likelihood w.r.t $\theta$. Using the prior as the proposal results in large discontinuities in the estimate when compared with the optimal proposal. Figures \ref{fig:KF_plots} (c) and (d) show that using CRN and the optimal proposal produces piece-wise continuous estimates. This is in contrast to using multinomial resampling where there a lot of fluctutations are apparent. We therefore advocate using an optimal proposal, or an approximation to such a proposal, in conjunction with CRN resampling. A good proposal will minimise the variance of the incremental weights at the current time-step. The process of selecting a good proposal can be time consuming but, as outlined in \cite{elfring2021particle}, can be critical in obtaining good results in other contexts.

We also note that, while optimisation algorithms are likely to be sensitive to discontinuities, we are focused on sampling $\theta$ and using MCMC to correct for the disparity between the proposal and the target.

\begin{algorithm}[]
\SetKwInput{KwInput}{Input} 
\DontPrintSemicolon
\KwInput{$\theta$, $y_{1:T}$}
\SetAlgoLined
Initialise: $x_0^i$, $\ddtheta x_0^i$, $\log(w_0^i)$, $\ddtheta \log(w_0^i)$\;
 \For{t = 1, \dots, T}{
  If necessary, resample $x_{t-1}^i$, $\log(w_{1:k-1}^i)$, $dx_{t-1}^i/d\theta$, $d\log(w_{1:t-1}^i)/d\theta$ as described in Section \ref{sect:resampling}.\;
  Sample the new particles $x_t^i$ and calculate the partial derivatives
			\begin{eqnarray}
				\partialdtheta f(x_{t-1}^i, \theta), \frac{\partial}{dx_{t-1}^i}f(x_{t-1}^i, \theta),
				\nonumber
			\end{eqnarray} from  (\ref{eqn:sampledeterministic}).\;
  Get the proposal mean, $\mu(x_{t-1}^i)$ and covariance, $C(x_{t-1}^i)$ for each particle $x_{t-1}^i$, as well as their derivatives
			\begin{eqnarray}
				\partialdtheta \mu(x_{t-1}^i), \frac{\partial}{x_{t-1}^i}\mu(x_{t-1}^i),
				\nonumber
			\end{eqnarray} 
			
			\begin{eqnarray}
				\partialdtheta C(x_{t-1}^i), \frac{\partial}{x_{t-1}^i}C(x_{t-1}^i),
				\nonumber
			\end{eqnarray} 
			
			seen in Section \ref{subsect:particlegradient}\label{op0}.\;
  Get the particle gradients $\ddtheta x_t^i$ using Subsection \ref{subsect:particlegradient}.\;
  Get the derivatives of the prior, proposal and likelihood using Subsections  \ref{subsect:dproposal}, \ref{subsect:dprior},  \ref{subsect:dlikelihood}.\;
  Evaluate the new log weights $\log w_{1:k}^i$ and log weight derivatives$\ddtheta \log w_{1:t}^i$ using  (\ref{eqn:weightupdate}) and (\ref{eqn:logweightderivative}), respectively.\;
  }
  Evaluate the final log likelihood, $\log p(y_{1:T} | \theta)$, and associated derivative, $\ddtheta \log p(y_{1:T} | \theta)$, using (\ref{eq:likelihood}) and (\ref{eqn:dloglik}), respectively.\;
 \caption{Particle Filter}
 \label{alg:the_alg_1}
\end{algorithm}

\section{Differentiable Particle Filters}\label{sec:diffPFs}

\subsubsection{Soft Resampling}
\label{sec:DP_soft}

{\itshape Soft resampling} was introduced in \cite{diffPF_4} and utilised in \cite{diffPF_1} and considers an approximation which involves drawing from the distribution $q(n)=\alpha w_{1:t}^{(\theta,i)}+(1-\alpha) 1 / N$, with $\alpha \in[0,1]$, representing a trade-off parameter. If $\alpha$ = 1, regular resampling is used and if $\alpha$ = 0 the algorithm performs subsampling. The new weights are calculated by 
\begin{eqnarray}
	w_{1:t}^{\prime(\theta,i)}=\frac{w_{1:t}^{(\theta,i)}}{\alpha w_{1:t}^{(\theta,i)}+(1-\alpha) 1 / N} \label{eqn:softresampling}.
\end{eqnarray}
This gives non-zero estimates of the gradient because the dependency on the previous weights is maintained. By changing $\alpha$, this method trades resampling quality for biased gradient estimates.

\subsubsection{Gumbel Softmax}
\label{sec:DP_gs}

The Gumbel-Max trick \cite{gumbel1954statistical} provides a way to sample a variable, $z$, from a categorical distribution that contains class probabilities, $\pi_x$. If we assume the categorical samples are one-hot vectors, $z$ can be sampled by

\begin{equation}\label{gumbel_softmax}
z=\text { onehot }\left(argmax_i\left\{G_{i}+\log \left(\pi_{i}\right)\right\}\right)
\end{equation}

\noindent where $x_i=G_{i}+\log(\pi_{i})$ and $G_i$ are independently sampled from $Gumbel(0,1)$. The summation in (\ref{gumbel_softmax}) is similar to the reparametrisation trick which is described previously however the {\itshape argmax} function is not differentiable. The work outlined in \cite{jang2016categorical} describes a differentiable approximation to {\itshape argmax} called {\itshape softmax} which is defined to be 

\begin{equation}
z =\frac{\exp \left(\frac{x_{k}}{\lambda}\right)}{\sum_{i=1}^{n} \exp \left(\frac{x_{i}}{\lambda}\right)}. 
\end{equation}

\noindent The temperature parameter, $\lambda$, is defined by the user and controls how closely the resulting Gumbel-softmax distribution approximates the categorical distribution. 

\subsubsection{Optimal Transport}
\label{sec:DP_ot}

A fully differentiable particle filter is described in \cite{diffPF_6} that resamples by using Optimal Transport ideas seen in \cite{peyre2020computational}. The log-likelihood estimate is asymptotically consistent but produces biased gradient estimates. The method needs additional hyper-parameters and runs in $O(N^2)$ time complexity so can be computationally expensive. 

\subsubsection{Fisher’s identity to calculate gradient of log-likelihood}
\label{sec:DP_FI}

As described in section \ref{sec:intro}, \cite{poyiadjis2011particle} recursively computes the gradient of the log-likelihood using Fisher's Identity, which can be summarised as follows:
\begin{align}
    \frac{d}{d\theta} \log p\left(y_{1:t}|\theta\right) =&   \frac{1}{p\left(y_{1:t}|\theta\right)}\frac{d}{d\theta} p\left(y_{1:t}|\theta\right) \\
    =&\frac{1}{p\left(y_{1:t}|\theta\right)}
    \frac{d}{d\theta}\int  p\left(y_{1:t},x_{1:t}|\theta\right)dx_{1:t}\\
    =&\frac{1}{p\left(y_{1:t}|\theta\right)}
    \int \frac{d}{d\theta} p\left(y_{1:t},x_{1:t}|\theta\right)dx_{1:t}\\
    =&
    \int \frac{p\left(y_{1:t},x_{1:t}|\theta\right)}{p\left(y_{1:t}|\theta\right)}\frac{d}{d\theta}\log p\left(y_{1:t},x_{1:t}|\theta\right)dx_{1:t}\\
=&
    \int p\left(x_{1:t}|y_{1:t},\theta\right)\frac{d}{d\theta}\log p\left(x_{1:t},y_{1:t}|\theta\right)dx_{1:t}\\
    \approx & \frac{1}{N}\sum_{i=1}^N \tilde{w}_{1:t}^{\left(\theta,i\right)} \underbrace{\frac{d}{d\theta}\log p\left(x^{\left(\theta,i\right)}_{1:t},y_{1:t}|\theta\right)}_{\alpha_{n}^{\thetai}}
\end{align}
Note that we can calculate the term inside the sum recursively as follows: 
\begin{equation}
\alpha_{n}^{\thetai}=\alpha_{n-1}^{\thetai}+ \ddtheta \log p\left(y_t | x_{t-1}^\thetai\right)+\ddtheta \log p\left(x_t^\thetai | x_{t-1}^\thetai, \theta\right),
\label{eq:unbiasedloglikelihoodestimate}\end{equation}


\noindent where 
$\log p\left(x_t^\thetai | x_{t-1}^\thetai, \theta\right)$ is the derivative of the prior and $\ddtheta \log p\left(y_t | x_{t-1}^\thetai\right)$ is the derivative of the log-likelihood (see (\ref{eq:incrementalweight}), (\ref{subsect:dprior}) and (\ref{subsect:dlikelihood}), respectively).

Recent work in \cite{diffresam_1} has outlined how to calculate this in the framework of Pytorch \cite{paszke2019pytorch} without having to modify the forward pass. They include a ``{\it{stop-gradient"}} operator that stops the gradients of the weights flowing into the resampling distribution. Much like the method described in this paper, minimal changes to the particle filtering algorithm need to be employed.

References \cite{poyiadjis2011particle} and \cite{del2015uniform} focus on having an unbiased estimator of the gradient of the log likelihood: we approximate with the gradient of the logarithm of the expectation whereas they calculate the expectation of the gradient of the logarithm. 
A feature of our approximation, perhaps surprisingly, is that it uses all samples from all iterations whereas the approach in \cite{poyiadjis2011particle} and \cite{del2015uniform} is recursively calculated along the trajectory of each particle. If, as is likely, the particle trajectories are degenerate, the approach in \cite{poyiadjis2011particle} and \cite{del2015uniform} will have a limited ability to use the diversity of states early in the trajectory to inform the gradient estimate. In short, we perceive that choosing between using the two approximations becomes a bias-variance trade-off.

\section{Estimation of parameters}
\label{sec:estimateparams}
If we have a prior, $p\left(\theta\right)$, for which the likelihood, $p\left(y_{1:T}|\theta\right)$, or log-likelihood\footnote{The log-likelihood is likely to be more stable numerically.} can be calculated, we can run p-MCMC to estimate $p\left(\theta|y_{1:T}\right)\propto p\left(\theta\right)p\left(y_{1:T}|\theta\right)$.
The gradient of the log-posterior of $\theta$ is given by
\begin{align}
\nabla \log p(\theta|y_{1:t})=\nabla \log p(\theta)+\nabla \log p(y_{1:t}|\theta),
\label{eq:gradientlogposterior}
\end{align}
where $\nabla \log p(\theta)$ is the gradient of the log-prior and $\nabla \log p(y_{1:t}|\theta)$ is the gradient of the log-likelihood. If we know $\nabla \log p(\theta|y_{1:t})$, it is possible to guide proposals to areas of higher probability within $\pi(\theta)$.

\subsection{Hamiltonian Monte Carlo}
HMC is a gradient based algorithm which uses Hamilton's equations to generate new proposals. Since it uses gradients, it is better at proposing  samples than a random-walk proposal. It was first developed in the late 1980s \cite{DUANE1987216} and in the last decade it has become a popular approach when implementing MCMC \cite{neal2012mcmc}. In the following section we give a high level conceptual overview of HMC and direct the reader to \cite{betancourt2017conceptual} for a more thorough explanation.
Hamilton's equations are a pair of differential equations that describe a system in terms of its position and momentum where the potential of the system is defined by $U=-\log(\pi(\theta))$. Conceptually, they are a set of differential equations that govern the movement of a frictionless puck across the potential surface. HMC introduces a momentum vector $m$  which moves the puck at $\theta$ on a random trajectory to $\theta'$. Intuitively, the puck will slow down or go backwards when going up a slope keeping the proposed $\theta$ in regions of higher probability. The total energy or Hamiltonian of a system can be expressed as
\begin{align}
H(\theta,m) = U(\theta) + K(m),
\end{align}
and is comprised of the sum of the Kinetic energy $K(m)$, which is dependent on where in the parameter space the puck is, and the potential energy $U(\theta)$, which is independent on the momentum $m$. 

Hamilton's equations describe how the system evolves as a function of time and are:
\begin{align}
\frac{d \theta}{d t}&=\frac{\partial H}{\partial m}\nonumber \\\\
\frac{d m}{d t}&=-\frac{\partial H}{\partial \theta}
\nonumber.
\end{align}
The joint density is
\begin{align}
p(\theta, m) \propto \exp(-H(\theta,m)) &= \exp(-U(\theta))\cdot\exp(-K(m)) \nonumber \\ &= p(\theta)p(m),
\end{align}
which means $\theta$ and $m$ are independent samples from the joint density so $m$ can be sampled from any distribution. For simplicity the distribution used is often chosen to be Gaussian and we make that choice here. 

Many numerical integration methods exist which discretise Hamilton’s equations and can be seen in \cite{hmcintegrators} with the leapfrog method being the go-to method for HMC. Leapfrog is a symplectic method which means the Hamiltonian remains close to its initial value, though not equal to it exactly, as the system is simulated. This means samples are generated with a high accept/reject ratio so the target is explored efficiently. Leapfrog is also a reversible method which allows detailed balance to be maintained. Finally, Leapfrog is a low-order method which uses relatively few gradient evaluations per step and is therefore computationally cheap. Algorithm \ref{alg:the_alg_1} is run every time a gradient evaluation is made within the Leapfrog numerical integrator. We note that the Pseudocode for Algorithm \ref{alg:the_alg_1} is specific to the problem described in section \ref{sec:discont} but can be easily applied to other models.

The samples generated are governed by a predetermined number of steps $L$ of size $\epsilon$, decided by the user. HMC is highly sensitive to the choice of these parameters, particularly $L$. If $L$ is too large, computation time can be wasted as the trajectory can make a U-turn and end close to where it started and, if too small, the proposal can exhibit random-walk behaviour. 

In some cases it has been shown that randomising $L$ can be beneficial to avoid periodicities in the underlying Hamiltonian dynamics \cite{bou2017randomized}. 

\subsection{No-U-Turn Sampler}
NUTS \cite{NUTS} is an extension of HMC and eliminates the need to specify $L$ by adaptively finding an optimal number of steps. NUTS does this by calculating

\begin{equation}\label{NUTS:criterion}
\frac{d}{d t} \frac{({\theta'}-\theta) \cdot({\theta'}-\theta)}{2}=({\theta'}-\theta) \cdot \frac{d}{d t}({\theta'}-\theta)=({\theta'}-\theta) \cdot \tilde{r}.
\end{equation}

\noindent and stopping the numerical integration when $({\theta'}-\theta) \cdot \tilde{r} < 0$. This indicates that performing any additional leapfrog steps would result in $\theta'$ not being proposed far enough away from $\theta$. To ensure (\ref{NUTS:criterion}) maintains {\itshape reversibility} a doubling method for slice sampling is undertaken. It is here a slice variable $u$ is introduced that has the conditional probability of
\begin{equation}\label{NUTS:sliceCondition}
p(u \mid \theta, r)= \text { Uniform }\left(u ;\left[0, \exp \left\{\mathcal{L}(\theta)-\frac{1}{2} r \cdot r\right\}\right]\right).
\end{equation}

\noindent After resampling $u \mid \theta, r$, NUTS builds a trajectory exploring states forwards and backwards in time by building a balanced tree, i.e. it takes forwards and backwards steps doubling in number each iteration. A Bernoulli trial is undertaken to decide the initial direction, and after the completion of the $j^{th}$ subtree a new trial is undertaken and a further $2^{j-1}$ steps are taken. This process continues until $\left(\theta^{+}-\theta^{-}\right) \cdot r^{-}<0 \quad \text { or } \quad\left(\theta^{+}-\theta^{-}\right) \cdot r^{+}<0$, which is similar to (\ref{NUTS:criterion}) but $(\theta^{-}, r^{-})$ and $(\theta^{+}, r^{+})$ is the left and right most leaves of the subtrees. This is when the trajectory begins to double back on itself (i.e. the puck starts to make a “U-Turn”) and  a state is sampled from a complete balanced tree. Two finite sets are introduced, $B$ and $C$, which include all of the nodes that were visited and all of the nodes that can be sampled that holds detailed balance, with $\mathcal{B} \supseteq \mathcal{C}$, respectively. All the candidate $(\theta, r)$ states in $C$ have to satisfy the slice condition in (\ref{NUTS:sliceCondition}). If one of these conditions is satisfied by a node or subtree from the last doubling iteration or the U-turn was made by the left- and right-most leaves of the balanced tree - detail balanced is maintained. The acceptance criteria differs from the standard Metropolis step as the final $\theta$ is sampled via the slice sample but other methods are presented in \cite{betancourt2016identifying}.

Another stopping criteria employed by NUTS is to stop the simulation when it is deemed to be not accurate enough. This is done by calculating 
\begin{equation}\label{NUTS_criteria}
-\mathcal{L}(\theta)-\frac{1}{2} r \cdot r-\log u<-\Delta_{\max },
\end{equation}
\noindent where $\Delta_{\max }$ is nonnegative which we set to be 1000. If (\ref{NUTS_criteria}) is satisfied then the simulation has reached a point of low probability, so any additional leapfrog steps are unlikely to be useful computations.

To avoid excessively large trees, which usually result from choosing too small a step-size, it is necessary to set a max-tree depth. To find a reasonable initial step-size we used the heuristic approach in \cite{NUTS} but we do not use dual averaging to automatically tune $\epsilon$. We anticipate doing so could improve performance. 

As the leap-frog step is volume-preserving, running HMC and NUTS can be run without computing the Jacobian determinants. Since using CRN in calculating the gradient from a particle filter results in gradients that are a deterministic function of the parameter, these Jacobian terms still cancel (and we do not need to integrate over the possible gradient calculations).

The pseudo-code for p-HMC and p-NUTS can be seen in Algorithm~\ref{alg:the_alg_2}. We implement the Algorithms 1 and 3 in \cite{NUTS} for HMC and NUTS respectively and note that each time a gradient evaluation is made during the leapfrog step, we run Algorithm \ref{alg:the_alg_1}.

\subsection{Metropolis-Adjusted Langevin Algorithm (MALA)}

MALA is a M-H proposal that includes gradient information about the log-posterior (as seen in \cite{Dahlin_2014}):

\begin{eqnarray}
    \label{MALA_eq}
	\theta^{\prime} = \mathrm{N}\left(\theta+\frac{1}{2} \Gamma \nabla \log p(\theta|y_{1:T}), \Gamma\right),
\end{eqnarray}

\noindent where $\Gamma=\gamma^{2} I_{d}$, and $\gamma$ is the step-size. Note that the step-sizes $\epsilon$ and $\gamma$ that are used in NUTS and $\Gamma$, as used in MALA, differ. In a similar way to \cite{Second-order}, we run the algorithm with different step-sizes and chose the one that provides an acceptance rate of around 0.3 in the stationary phase. Although we have presented MALA as an independent proposal it is a special case of HMC when $L=1$. However, they both have different acceptance steps.

\begin{algorithm}[]
\SetKwFunction{FMain}{Leapfrog}
\SetKwInput{KwInput}{Input} 
\DontPrintSemicolon

\SetKwProg{Fn}{Function}{:}{}

\KwInput{$\theta_{0}$, $y_{1:T}$, $M$, $L$}
\SetAlgoLined
 $\ell(\theta)$, $\nabla\mathcal{L}(\theta)$ = Run Algorithm \ref{alg:the_alg_1}\; 
 $\epsilon$ = Find Reasonable $\epsilon$($\theta_{0}$)\;
 \For{i = 1 to M}{
 
 HMC = Algorithm 1 in  \cite{NUTS} or \\
 
 NUTS = Algorithm 3 in \cite{NUTS}
 
 }
 \
 
 \Fn{\FMain{$\theta$, $m$, $y_{1:T}$, $\nabla\mathcal{L}(\theta)$}}{
        $m'$ = $m$ + 0.5 $\cdot$ $\epsilon$ $\cdot$ $\nabla\mathcal{L}(\theta)$\;
        $\theta'$ = $\theta$ + $\epsilon$ $\cdot$ $m'$\;
        $\ell(\theta)$, $\nabla\mathcal{L}(\theta)'$ = Run Algorithm \ref{alg:the_alg_1}\;
        $m'$ = $m'$ + 0.5 $\cdot$ $\epsilon$ $\cdot$ $\nabla\mathcal{L}(\theta)'$
        
        \KwRet $\theta'$, $m'$, $\ell(\theta)'$, $\nabla\mathcal{L}(\theta)'$\;
  }

 \caption{Particle - HMC or NUTS}
 \label{alg:the_alg_2}
\end{algorithm}

\section{Numerical Experiments}
\label{sec:testingestimates}
\subsection{Linear Gaussian State Space Model}
We consider the Linear Gaussian State Space (LGSS) model seen in Section 4 of \cite{dahlin2015getting} which is given by
\begin{equation}
x_{t} \mid x_{t-1} \sim \mathcal{N}\left(x_{t} ; \phi x_{t-1}, \sigma_{v}^{2}\right),
\label{LGSSXt}
\end{equation}
\begin{equation}
y_{t} \mid x_{t} \sim \mathcal{N}\left(y_{t} ; x_{t}, \sigma_{e}^{2}\right),
\label{LGSSYt}
\end{equation}
where $\theta=\left\{\phi, \sigma_{v}, \sigma_{e}\right\}$ are parameters with prior densities Normal$(0,1)$, Gamma$(1,1)$ and Gamma$(1,1)$, respectively. The ``optimal'' proposal is used (see (\ref{eq:optimalproposalpropogate})) and can be derived from the properties of (\ref{LGSSXt}) and (\ref{LGSSYt}). This results in
\begin{equation}
q\left(x_t|x_{t-1},y_t\right) =\mathcal{N}\left(x_{t} ; \sigma^{2}\left[\sigma_{e}^{-2} y_{t}+\sigma_{v}^{-2} \phi x_{t-1}\right], \sigma^{2}\right),
\end{equation}
with $\sigma^{-2}=\sigma_{v}^{-2}+\sigma_{e}^{-2}$.

The weights are updated using (\ref{eq:optimalproposalweight}), which can be shown to be\\
\begin{equation}
w_{1:t}^{(\theta,i)}=\mathcal{N}\left(y_{t} ; \phi x_{t}, \sigma_{v}^{2}+\sigma_{e}^{2}\right)w_{1:t-1}^{(\theta,i)}.
\end{equation}

\subsubsection{Results}

\begin{landscape}
\begin{table*}[t!]
\centering
\begin{tabular}{c|ccccc|ccccc|ccccc} 
  & \multicolumn{5}{c|}{\textbf{T=25}} & \multicolumn{5}{c|}{\textbf{T=50}} &  \multicolumn{5}{c}{\textbf{T=100}}  \\

 & CRN & SR & GS & OT & FI & CRN & SR & GS & OT & FI & CRN & SR & GS & OT & FI \\\hline \hline
\bf{N=16}  & &  & &  &  &  &  & & & &  &  &  &  & \\
 MSE & 0.151 & 1.446 & 0.179 & 0.141 & \bf{0.135} & 0.075 & 0.438 & 0.076 & \bf{0.025} &\bf{0.025} & \bf{0.035} & 0.056 & 0.045 & 0.040 & 0.047\\
 Time (sec) & 593 & 730 & 554 & 2062 & 582 & 728 & 1423 & 765 &6256 &1590 & 939 & 2684  & 1018 & 4396 & 1005\\ \hline \hline
\bf{N=32}  & &  & &  &  &  &  & & & &  &  &  &  & \\
 MSE & \bf{0.112} & 0.899 & 0.174 & 0.124 & 0.121 & 0.077 & 0.297 & 0.078 & \bf{0.020}  & 0.021 & \bf{0.035} & 0.050  & 0.045 & \bf{0.035} & 0.036\\
 Time (sec) & 558 & 816 & 587 & 1667 & 589 & 790 & 1478 & 784 &  6839 & 1339&941 & 2601 & 1583 & 5292 & 1014 \\ \hline \hline
\bf{N=64}  & &  & &  &  &  &  & & & &  &  &  &  & \\
 MSE & \bf{0.121} & 1.091 & 0.165 & \bf{0.121} &0.126 & 0.074 & 0.452 & 0.084 &0.069 & \bf{0.025} & \bf{0.025} & 0.396 & 0.050 & 0.044 & 0.049\\
 Time (sec) & 552 & 747 & 852 & 1647 & 608& 750 & 1481 & 1229 & 2303 &1369 & 1952 & 2571 & 2959 & 5281 & 1022\\ \hline \hline
\bf{N=128}  & &  & &  &  &  &  & & & &  &  &  &  & \\
 MSE & \bf{0.129} & 1.576 & 0.135 & 0.142 &0.162 & 0.078 & 0.318 & 0.074 & 0.082 &\bf{0.027} & \bf{0.026} & 0.825 & 0.043 & 0.044 & 0.044\\
 Time (sec) & 583 & 880 & 1152 & 1827 &582 & 738 & 1452 & 1423 & 2592 &1383& 1586 & 2621 & 3034 & 5651 & 1020 

\end{tabular}
\caption{Time in seconds and the average MSE of $\theta=\left\{\phi, \sigma_{v}\right\}$ in the LGSS model for different numbers of $N$, $T$ and differentiable particle filters outlined in section \ref{sec:diffPFs}. CRN = us, SR = soft resampling (see section \ref{sec:DP_soft}), GS = gumbel-softmax (see section \ref{sec:DP_gs}), OT = optimal transport (see section \ref{sec:DP_ot}) and FI = Fisher’s identity (see section \ref{sec:DP_FI}). The results are an average over 10 runs using different random number seeds and NUTS was used as the proposal.}
\label{table:diffPF}
\normalsize
\end{table*}
\end{landscape}

\begin{table}[]
\centering
\scriptsize
\begin{tabular}{cc|ccc|ccc}
  & & \multicolumn{3}{c|}{\textbf{N = 512}} & \multicolumn{3}{c}{\textbf{N = 1024}}   \\
  & & \textit{T (Secs)}  & \textit{MSE} & \textit{NGE} & \textit{T (Secs)} & \textit{MSE} & \textit{NGE} \\
   \hline
   \hline
& MALA  &  4.342 &   0.306         & 9   &  6.288 &  0.300         & 9 \\ 
   \hline
   \hline
{\ul \parbox[t]{2mm}{\multirow{10}{*}{\rotatebox[origin=c]{90}{\textbf{HMC}}}}}
& L1   &   1.83  &   0.39$\pm$0.01 & 9   &  8.06  &  0.40$\pm$0.01 & 9\\      
& L2   &   3.70  &   0.36$\pm$0.01 & 18  &  14.21 &  0.36$\pm$0.02 & 18\\  
& L3   &   5.61  &   0.34$\pm$0.02 & 27  &  21.38 &  0.33$\pm$0.03 & 27\\  
& L4   &   7.26  &   0.31$\pm$0.02 & 36  &  32.34 &  0.31$\pm$0.04 & 36\\  
& L5   &   7.83  &   0.30$\pm$0.02 & 45  &  46.31 &  0.29$\pm$0.05 & 45\\  
& L6   &   11.39 &   0.27$\pm$0.03 & 54  &  57.84 &  0.26$\pm$0.06 & 54\\  
& L7   &   14.43 &   0.26$\pm$0.04 & 63  &  47.67 &  0.29$\pm$0.07 & 63\\  
& L8   &   18.86 &   0.25$\pm$0.05 & 72  &  54.18 &  0.25$\pm$0.07 & 72\\  
& L9   &   22.75 &   0.24$\pm$0.06 & 81  &  58.54 &  0.23$\pm$0.08 & 81\\  
&L10  &   24.46 &   0.22$\pm$0.06 & 90  &  103.80&  0.23$\pm$0.08 & 90\\  
\hline
\hline
&NUTS & 35.77    &   0.23$\pm$0.08  & 106 &  69.91 &   0.19$\pm$0.07  & 66\\ 
\end{tabular}
\caption{The time taken in seconds, number of gradient evaluations (NGE) and MSE and standard deviation of $\theta=\left\{\phi, \sigma_{v}, \sigma_{e}\right\}$ in the LGSS model for different numbers of $N$ and MCMC proposals. $T=100$ observations, $M=10$ MCMC iterations and CRN resampling was used. The results were averaged over 10 runs using different random number seeds.}
\normalsize
\label{table:comparingLGSSM}
\end{table}

First, we compare the different differentiable particle filters described in section \ref{sec:diffPFs} in terms of computational run-time and the MSE between the true and inferred values of $\theta=\left\{\phi, \sigma_{v}\right\}$ in the LGSSM described in (\ref{LGSSXt}) and (\ref{LGSSYt}). The true values of $\phi$ and $\sigma_{v}$ are 0.7 and 1.2 respectively. We use NUTS as the proposal and run with different numbers of particles, $N$, and observations, $T$, over $M=50$ MCMC iterations. The results are presented in Table \ref{table:diffPF} with the MSE and the time taken in seconds being averages over 10 runs (with different random number seeds). 

Using CRN and FI (Fisher's identity) consistently results in the lowest time taken to complete the different experiments. One reason for this is that minimal changes to the resampling step are introduced when compared to the other methods in Table \ref{table:diffPF}. When $T=25$ and $T=100$, using CRN results in the lowest MSE when running with $N=32$, 64 and 128 and $N=16$, 32, 64, and 128, respectively. Using optimal transport resampling results in lower MSE estimates in some experiments but the computation time is considerably higher than CRN and FI.

Next, we compare the different proposals outlined in section \ref{sec:estimateparams}. It has been explained previously that using NUTS eliminates the need to manually select the length parameter, $L$, in HMC by adaptively choosing this parameter at every iteration. Therefore it is likely that NUTS will make more than one target evaluation per iteration so is more computationally costly than MALA, where only one evaluation is made, and HMC with certain values for $L$. We therefore include the number of gradient evaluations as well as the average MSE between the true and inferred values of $\theta=\left\{\phi, \sigma_{v}, \sigma_{e}\right\}$ and the computation time in seconds for the LGSSM described in (\ref{LGSSXt}) and (\ref{LGSSYt}). The true values of $\phi$, $\sigma_{v}$ and $\sigma_{e}$ are 0.7, 1.2 and 1, respectively. The setup of the experiment was as follows: $T=100$, $M=10$, $N=512$ and 1024 and CRN resampling was used. Table \ref{table:comparingLGSSM} exemplifies the benefit of NUTS over HMC in not having to optimise the number of steps, $L$: NUTS  obtains a lower MSE in a shorter run-time than HMC.

\begin{figure}
\centering
\begin{subfigure}[b]{\textwidth}
\begin{center}
\includegraphics[width=\linewidth]{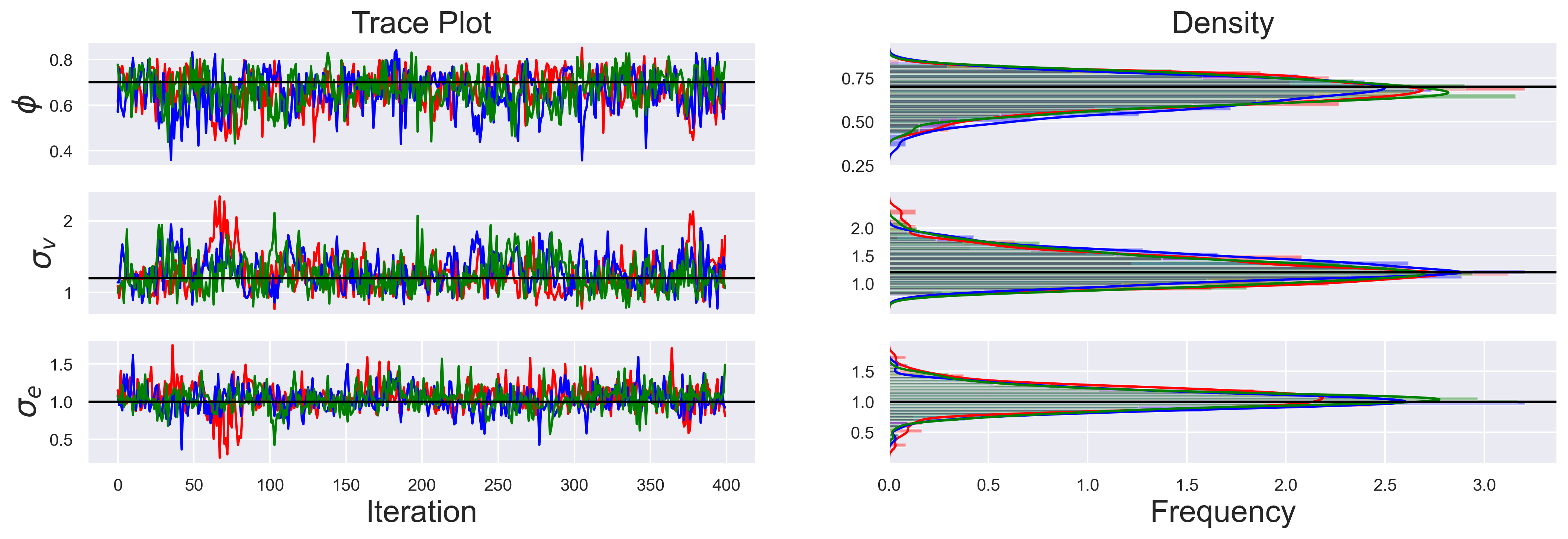}
\caption{}
\end{center}
\end{subfigure}
\hfill
\begin{subfigure}[b]{0.45\textwidth}
\includegraphics[width=\linewidth]{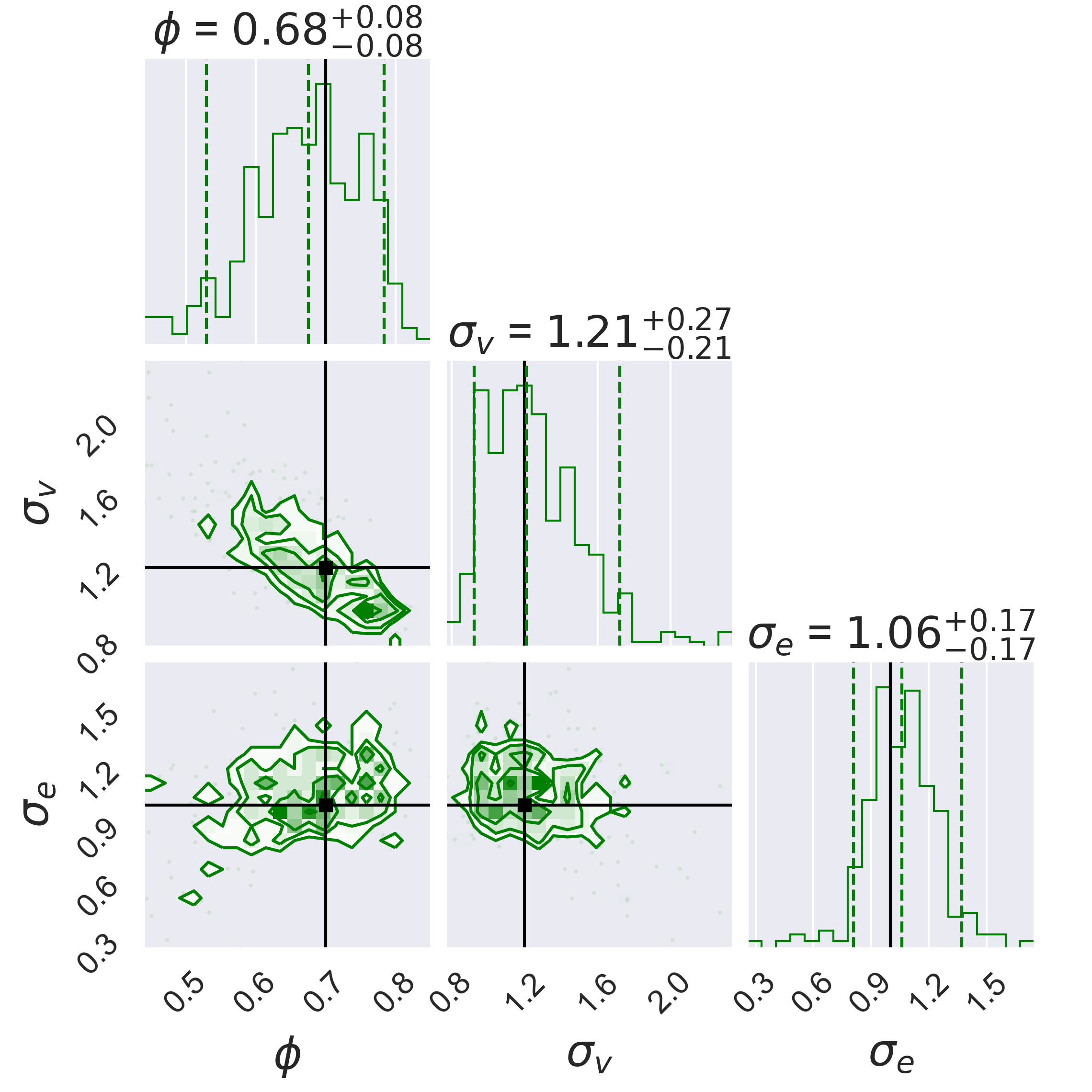}
\caption{}
\end{subfigure}
\hfill
\centering
\begin{subfigure}[b]{0.45\textwidth}
\includegraphics[width=\linewidth]{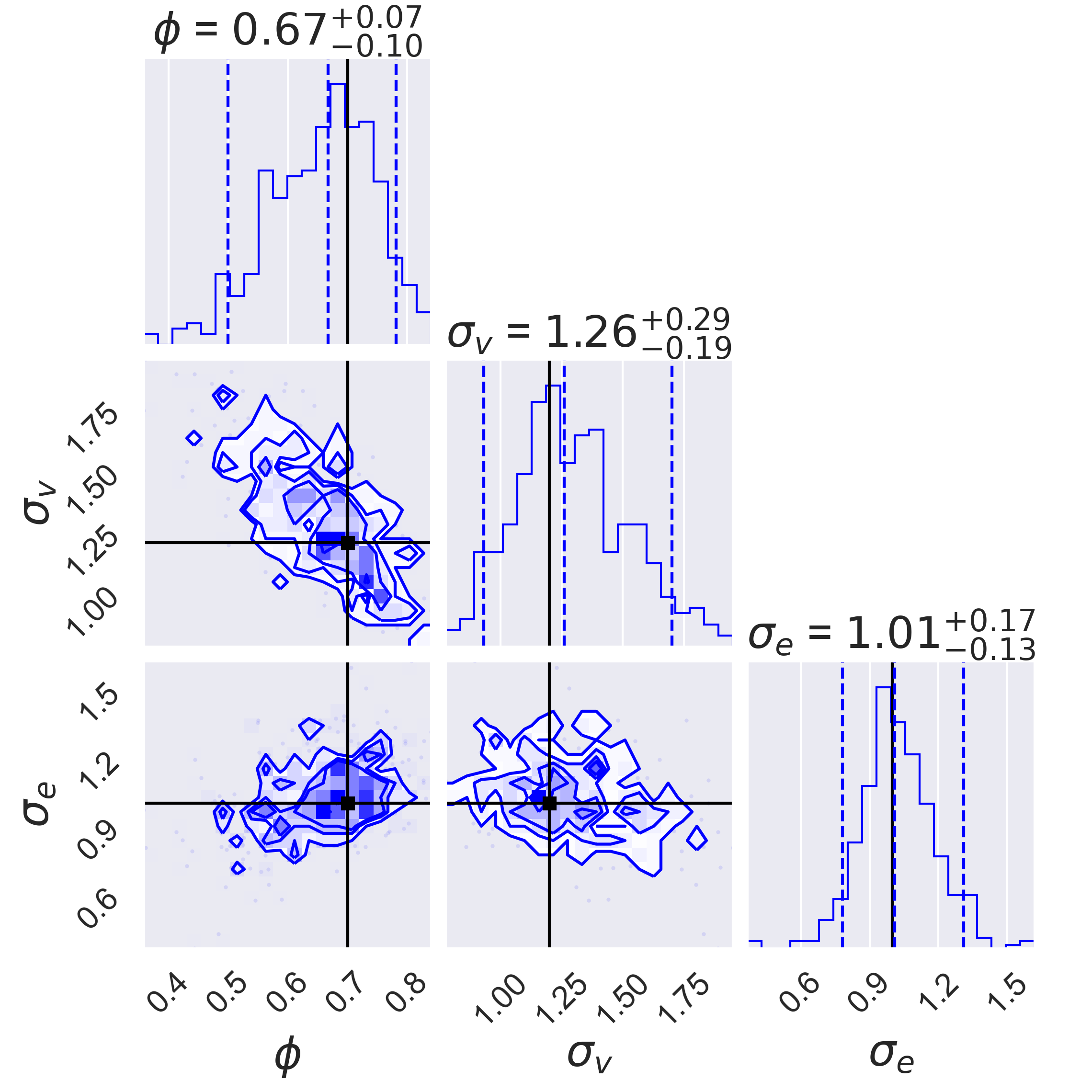}
\caption{}
\end{subfigure}
\hfill
\centering
\begin{subfigure}[b]{0.45\textwidth}
\includegraphics[width=\linewidth]{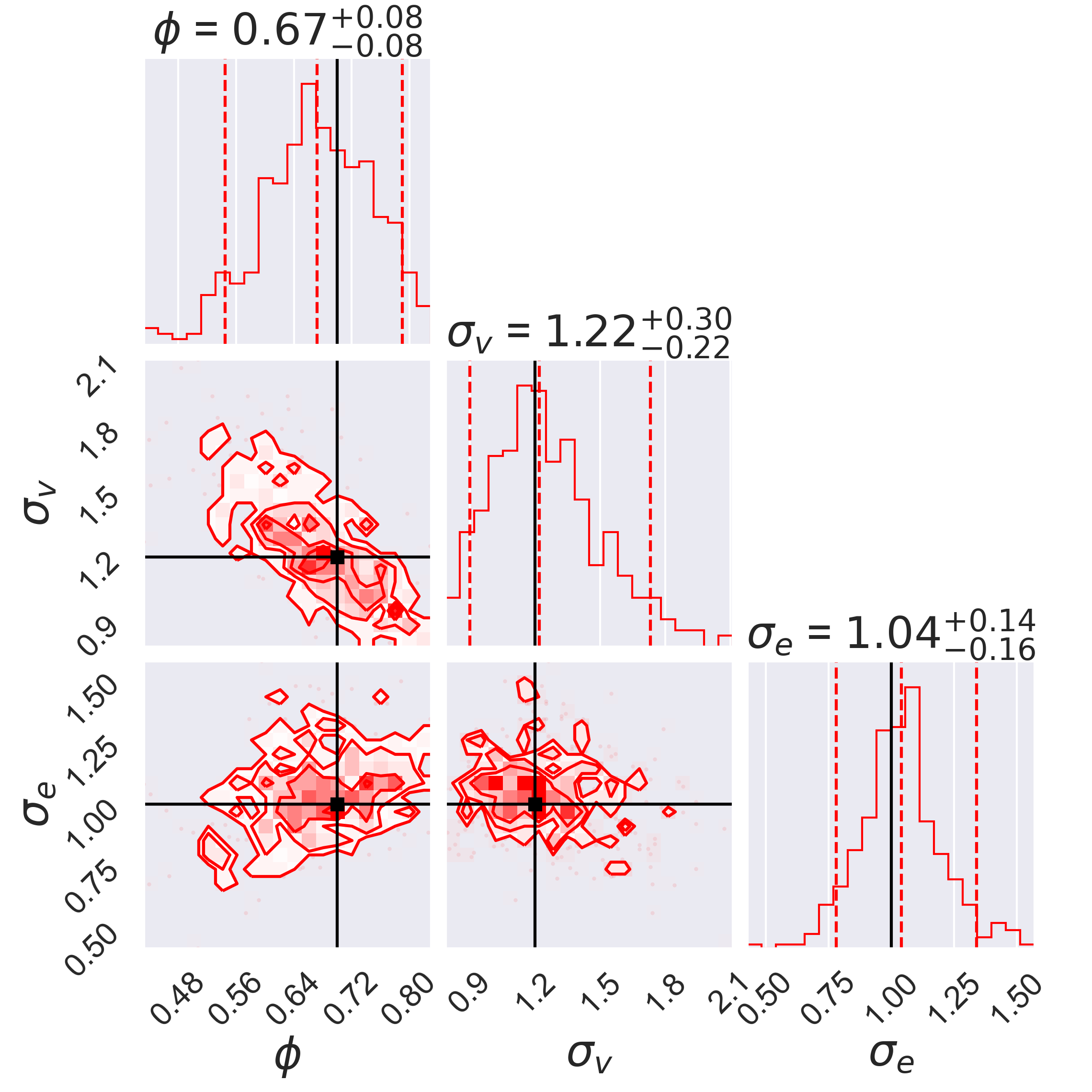}
\caption{}
\end{subfigure}
\caption{(a) The trace plots (left) and density estimates (right) of three independent chains with diﬀerent initial starting positions for $\theta$. The horizontal black lines indicate the true values. (b)-(d) 1-Dimensional corner plots of the three independent chains seen in Figure 2 for the LGSS model. The black lines indicate the true values.}
\label{fig:traceplots_LGSSM}
\end{figure}

Finally, we show results when using NUTS and CRN resampling and present a commonly used diagnostic to determine if three independent chains that had different initial starting values for $\theta=\left\{\phi, \sigma_{v}, \sigma_{e}\right\}$ have converged. The particle filter was initialised with $N=750$, $T=250$ observations, $M=500$ (with the first 100 discarded as burn-in) and the true values were $\theta=[0.7, 1.2, 1]$. The trace plots and density plots of the accepted samples of $\theta$ can be seen on the left and right of Figure \ref{fig:traceplots_LGSSM} (a), respectively. These plots give an indication of how well the chains have converged to their stationary distribution. Figure \ref{fig:traceplots_LGSSM}(b)-(d) shows the 1-dimensional histograms, plotted using \cite{cornerPlots_python}, for the same three chains and the uncertainties associated with these estimates of $\theta$. The mean estimate of $\theta$ from the three chains is $[0.67, 1.23, 1.04]$. A numerical method for determining if multiple chains have converged is the Gelman-Rubin diagnostic \cite{gelman_rubin} which compares the variances between chains. It is a commonly used diagnostic in the probabilistic programming language Stan \cite{stan} (where it is referred to as $\hat{R}$) to ascertain if the sampler has correctly sampled from the posterior. Stan's documentation states an $\hat{R}$ value below 1.05 passes their internal diagnostic check. Table \ref{table:gelman_rubin} shows the calculated $\hat{R}$ values for $\theta$, which are all below 1.05.

\begin{table}[]
\centering
\begin{tabular}{c | c | c | c }
& $\mu$ & $\phi$ & $\sigma_v$ \\
\hline \hline
Gelman-Rubin & 1.0091 & 1.007 & 1.0094 \\ 
\end{tabular}
\caption{The Gelman-Rubin statistic for each dimension of $\theta$ in the LGSS model. The results correspond to the three chains presented in Figure \ref{fig:traceplots_LGSSM}.}
\label{table:gelman_rubin}
\end{table}

\subsection{Stochastic Volatility Model}
\label{sec:svm}
Stochastic volatility (SV) models are widely used to evaluate financial securities and prices \cite{hullwhitefinance}, as the variance of the latent process changes over time and is not constant. This is modelled with the state $x_t$ being proportional to the observation noise. More specifically, we use the same model as seen in~\cite{dahlin2015getting}, which is defined as follows:
\begin{equation}
\label{SVmodel1}
x_{0} \sim \mathcal{N}\left(x_{0} ; \mu, \frac{\sigma_{v}^{2}}{1-\phi^{2}}\right),
\end{equation}

\begin{equation}
\label{SVmodel2}
x_{t+1} \mid x_{t} \sim \mathcal{N}\left(x_{t+1} ; \mu+\phi\left(x_{t}-\mu\right), \sigma_{v}^{2}\right),
\end{equation}

\begin{equation}
\label{SVmodel3}
y_{t} \mid x_{t} \sim \mathcal{N}\left(y_{t} ; 0, \exp \left(x_{t}\right)\right),
\end{equation}

\noindent where $\theta=[\mu,\phi,\sigma_{v}]$ are parameters with prior densities Normal$(0,1)$, Normal$(0,1)$ and Gamma$(2,10)$, respectively. The log-returns (observations), $y_t$, are modelled using the formula,
\begin{equation}
y_{t}=100 \log \left[\frac{s_{t}}{s_{t-1}}\right]=100\left[\log \left(s_{t}\right)-\log \left(s_{t-1}\right)\right]
\end{equation}
where $s_t$ is the price. We note that the data used here is real data and is the daily closing prices of the NASDAQ OMXS30 index, i.e., a weighted average of the 30 most traded stocks at the Stockholm stock exchange. The data is extracted from Quandl\footnote{The data can be downloaded from \url{https://data.nasdaq.com/data/NASDAQOMX/OMXS30}} for the period between January 2, 2012 and January 2, 2014.

We use the prior as the proposal (see (\ref{eq:priorproposal})), which is such that the incremental log weight becomes
\begin{eqnarray}
	\log \sigma\left(x_t^\thetai, x_{t-1}^\thetai, \theta\right) = \log p\left(y_t | x_{t}^\thetai\right),
\end{eqnarray}
and the associated gradient is
\begin{eqnarray}
	\frac{d}{d \theta} \log \sigma\left(x_{t}^{(\theta, i)}, x_{t-1}^{(\theta, i)}, \theta\right)=\frac{d}{d \theta} \log p\left(y_{t} \mid x_{t-1}^{(\theta, i)}\right).
\end{eqnarray}

\subsubsection{Results}

It is here we compare the different proposals outlined in section \ref{sec:estimateparams} when inferring the parameters of the stochastic volatility model outlined in (\ref{SVmodel1}) - (\ref{SVmodel3}). Each simulation was initialised with the same values of $\theta$, $N=5000$, $T=500$ observations, $M=5000$ MCMC iterations with the first 2000 discarded as burn-in and CRN resampling. 

\begin{figure*}[]
\centering
    \includegraphics[width=1\textwidth]{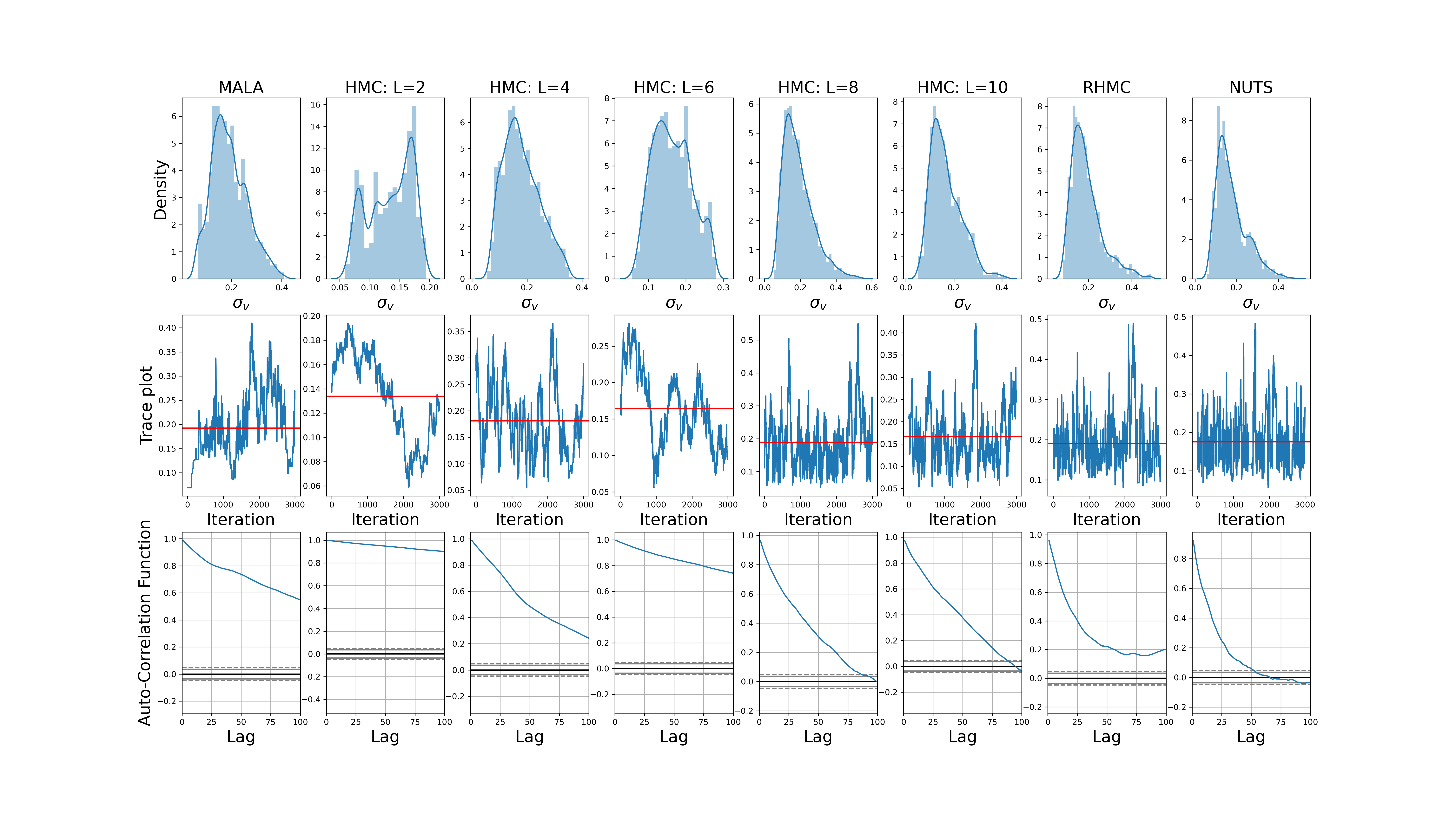}
    \caption{Columns: Simulations using MALA, HMC with different $L$, RHMC and NUTS. First row: Histograms of posterior estimate of $\sigma_{v}$. Second row: Trace plots of $\sigma_{v}$ and the red horizontal line is the estimated mean. Third row: ACF plots for $\sigma_{v}$ with lag = 100.}
    \label{fig:NUTSvsHMC}
\end{figure*}

The first and second rows of figure \ref{fig:NUTSvsHMC} shows the histograms and trace plots of the accepted values of $\sigma_v$, respectively. These plots give a visual indication of how well each of the samplers perform but should not be solely used to asses convergence. Table \ref{table:SV_results} outlines a number of MCMC diagnostics that determine if the sampler has converged to equilibrium. They include the mean of the posterior samples which on its own is not very informative, especially if the parameter being inferred is not known. 

By looking at the trace plots, it is evident that for MALA and some values of $L$ in HMC, there is a lot of serial correlation between consecutive draws. This results in the parameter space being poorly explored. However using NUTS can be seen to have the least serial correlation between MCMC draws. This observation is backed up by looking at the third row, which shows auto-correlation function (ACF) plots for the same simulations. These plots show the auto-correlation for a Markov chain up to a user-specified number of lags, which in this case is chosen to be 100. An ideal ACF plot is large at short lags but quickly drops towards zero. For MALA and a number of the HMC simulations the ACF plots do not reach 0 within the specified 100 lags. The Integrated Auto-Correlation Time (IACT) is a measure of the area under the ACF plot. The aim is to minimise this value since it gives an indication of the mixing within the Markov chain: IACT estimates the number of iterations it takes to draw an independent sample. It is evident by looking at Table \ref{table:SV_results} that using NUTS results in lower IACT scores for parameters $\mu$ and $\sigma_v$ and is comparable with HMC with $L=6$ for $\phi$.

The effective sample size (ESS) is also shown in Table \ref{table:SV_results}. This gives an indication of the number of independent samples it would take to have the same estimation power as a set of auto-correlated samples: larger values for ESS are to be preferred. Much like the IACT scores for parameters $\mu$ and $\sigma_v$, NUTS provides better results than the other samplers and is slightly worse than HMC with $L=6$ for $\phi$. 

As explained previously, randomising the $L$ parameter within HMC can avoid periodicities in the underlying Hamiltonian dynamics. To do this we draw an $L$ value from an exponential distribution with a mean parameter of 2.5. To ensure this value is an integer we round up. It is evident when looking at Table \ref{table:SV_results}, randomising the $L$ parameter in HMC provides better results for $\mu$ when compared with HMC with fixed $L$ but is still worse than NUTS.

The mean estimates of $\theta$ differ slightly in each simulation to those presented in \cite{dahlin2015getting}, which were $[-0.23, 0.97, 0.15]$. We believe this disparity stems from \cite{dahlin2015getting} using the same particle filter but a M-H random walk proposal for the parameters. However, when a reparameterised model (described on page 29 of \cite{dahlin2015getting}) is used, they obtain estimates equal to $[-0.16, 0.96, 0.17]$, which are very similar to those seen in Table \ref{table:SV_results} when using NUTS, $[-0.17, 0.96, 0.18]$.

\begin{table*}[t!]
\tiny
\centering
\begin{tabular}{cc|c|cccccccccc|c|c} 

  & & {\textbf{MALA}} & \multicolumn{9}{c}{\textbf{HMC}} & &{\textbf{RHMC}}         & {\textbf{NUTS}} \\
 & &  & L1  & L2   & L3    & L4    & L5    & L6   & L7    & L8    & L9   & L10   &     &                                  \\ \hline \hline
{\ul \parbox[t]{2mm}{\multirow{3}{*}{\rotatebox[origin=c]{90}{\textbf{Mean}}}}}   
& $\mu$                               & -0.27 & -0.12              & 0.01 & -0.34 & -0.01 & -0.16 & -0.20 & -0.24 & -0.16 & -0.30 & -0.17 & -0.16  & -0.17                              \\
& $\phi$                             & 0.95 & 0.97               & 0.94 & 0.96  & 0.97  & 0.96  & 0.96 & 0.95  & 0.95  & 0.96 & 0.93  & 0.96   &  0.96                               \\
& $\sigma_{v}$                  &  0.19  & 0.13               & 0.23 & 0.18  & 0.14  & 0.16  & 0.18 & 0.19  & 0.2   & 0.17 & 0.23  & 0.19   & 0.18                               \\ \hline \hline
{\ul \parbox[t]{2mm}{\multirow{3}{*}{\rotatebox[origin=c]{90}{\textbf{IACT}}}}}  
& $\mu$                              & 131  & 173                & 190  & 182   & 164   & 192   & 154  & 125   & 163   & 167  & 174   & 101   &   66                              \\
& $\phi$                           & 125    & 158                & 175  & 100   & 98    & 139   & 35   & 68    & 79    & 73   & 144   & 63    &   36                             \\
& $\sigma_{v}$                       & 148  & 191                & 181  & 111   & 137   & 173   & 44   & 74    & 89    & 81   & 146   & 66   &  35                               \\ \hline \hline
{\ul \parbox[t]{2mm}{\multirow{3}{*}{\rotatebox[origin=c]{90}{\textbf{ESS}}}}} 
& $\mu$                              & 15  & 5                & 1  & 2   & 8   & 1   & 12  & 16  & 4   & 11  & 1  & 30   & 50                                \\
& $\phi$                           & 3   & 2  & 2  & 25   & 15    & 8  & 94   & 59   & 37    & 42   & 18   & 75     &  89                             \\
& $\sigma_{v}$                       & 2  & 1  & 2  & 26   & 6   & 6   & 71   & 50    & 34    & 44   & 16   & 55  &  102                              \\ \hline \hline
 & {\ul \textbf{Acc. rate}} & 0.35  & 0.73               & 0.81 & 0.73  & 0.87  & 0.83  & 0.70  & 0.66  & 0.76  & 0.68 & 0.70   &  0.53  &  0.86 

\end{tabular}
\caption{IACT, ESS and mean estimates of $\theta$ in the SV model and the acceptance probability of the different algorithms. The same value of $\epsilon$ was used for each simulation of HMC, RHMC and NUTS, $N=1000$, $M=5000$ and CRN resampling was used.}
\label{table:SV_results}
\normalsize
\end{table*}

\subsection{Lorenz-63}
The Lorenz-63 model is a 3-dimensional ($N_{x}$) dynamical system widely used in data assimilation \cite{cocucci2021model} that uses ordinary differential equations to propagate a state. The model has a state variable which is propagated with:


\begin{equation}
\label{L63_1}
\frac{dx}{dt} = \sigma(y-x),
\end{equation}

\begin{equation}
\label{L63_2}
\frac{dy}{dt} = x(r-z)-y,
\end{equation}

\begin{equation}
\label{L63_3}
\frac{dz}{dt} = xy-bz,
\end{equation}


\noindent where $ \sigma=10, r=28, b=8/3$, which represent the Prandtl number, the Rayleigh number and the physical dimensions of the layer respectively. This model is commonly used in meteorology and oceanography due to its non-linear chaotic behaviour. The assimilation cycle length is $dt=0.05$ time units in every experiment. This corresponds to 50 integration time steps which are performed with a 4th-order Runge-Kutta algorithm.

We are interested in the following model:
\begin{equation}
x_{t} \mid x_{t-1} \sim \mathcal{N}(x_{t} ; \mathcal{M}(x_{t-1}), \mathcal{Q})
\end{equation}
\begin{equation}
y_{t} \mid x_{t} \sim \mathcal{N}(y_{t} ;\mathcal{H}x_{t} , \mathcal{R})
\end{equation}

\noindent where $\mathcal{M}$ is the Lorenz propagation and $\mathcal{H}$ is the observational model. The parameters in the model and their prior densities are $\theta=\{\mathcal{Q},\mathcal{R} \}$ and Normal$(0,1)$ and Normal$(0,1)$, respectively.

The observational model, $\mathcal{H}$, is a $N_y \times N_x$ matrix where $N_{y}$ represents the dimension of the observations with $N_{y} < N_{x}$. The motivation for having a partially observable model is to evaluate our proposed method in a realistic environment. We use the prior as the proposal as described in (\ref{eq:priorproposal}).

The true values of the parameters are as follows: $N_x = 3$, $N_x = 2$, $\mathcal{Q} = \sigma_{\mathcal{Q}} \mathbb{I}_{N_x} $ and $\mathcal{R} = \sigma_{\mathcal{R}} \mathbb{I}_{N_y} $ where $\mathbb{I}_{N_x}$ and $\mathbb{I}_{N_y}$ are $N_x \times N_x$ and $N_y \times N_y$ identity matrices, respectively. $\sigma_{\mathcal{Q}}$ and $\sigma_{\mathcal{R}}$ are both set to 1.2.

\subsubsection{Results}

For this example we compare HMC with different $L$ and NUTS when using the different differentiable particle filters described in section \ref{sec:diffPFs} in terms of the computational run-time and the MSE between the true and inferred values of $\theta = {\sigma_{\mathcal{Q}}}$. Note we do not include results for MALA and RHMC proposals because we make use of the HMC and NUTS implementations provided by PyTorch \cite{paszke2019pytorch}.

Each simulation was initialised with the same values of $\theta$, $N=500$, $T=50$ observations, $M=10$ MCMC iterations using CRN, Gumbel-Softmax resampling (GS) and Soft-resampling (SR). Note we do not include results for OT resampling due to the computation time being excessive: it takes 3.01 seconds for 1 gradient evaluation with CRN compared with 22.76 seconds with OT. We also note that running the experiments on a GPU did not significantly reduce the computation time. 

It is evident when looking at Table \ref{table:Lorenz_results} that using NUTS and CRN resampling yields the lowest MSE. One explanation for  Gumbel-softmax and soft-resampling giving worse results than CRN is because these functions are approximations of multinomial-resampling which is unbiased. We note that fine tuning $\alpha$ and $\lambda$ in (\ref{eqn:softresampling}) and (\ref{gumbel_softmax}), respectively may result in more accurate estimates of $\theta$.

\begin{table}[]
\scriptsize
\centering
\begin{tabular}{ll|lllll|l} 

  & &  \multicolumn{5}{c}{\textbf{HMC}} &  \textbf{NUTS}  \\
  & &  L2     & L4        & L6      & L8       & L10   &                                \\ \hline \hline
{\ul \parbox[t]{2mm}{\multirow{2}{*}{\rotatebox[origin=c]{90}{\textbf{CRN}}}}} 
 & MSE       & 0.42 & 0.40  & 0.19  & 0.17  & 0.18  & 0.12  \\ 
 & Time (s)  & 98   & 360   & 1458  & 4860  & 5250  & 1560  \\ \hline \hline

{\ul \parbox[t]{2mm}{\multirow{2}{*}{\rotatebox[origin=c]{90}{\textbf{GS}}}}}                  
 & MSE        & 0.36 & 0.51  & 0.59  & 0.59 & 0.21      & 0.21  \\
 & Time (s)   & 83   & 644   & 1900  & 6600 &  12900     & 1255  \\
 \hline \hline

{\ul \parbox[t]{2mm}{\multirow{2}{*}{\rotatebox[origin=c]{90}{\textbf{SR}}}}}                  
 & MSE        & 0.39 & 0.33  & 1.2  & 3.5 &   0.67    & 0.20  \\
 & Time (s)   & 119   & 617   & 2141  & 9180 &  52200     & 869  \\

\end{tabular}
\caption{Mean square errors (for the $\sigma_{\mathcal{Q}}$ parameter) and computational times for different MCMC proposals and resampling schemes. The same value of $\epsilon$ was used for each simulation of HMC and NUTS, $N=500$, $M=10$, $T=50$}
\label{table:Lorenz_results}
\normalsize
\end{table}

\section{Conclusion}
\label{sec:conclusion}
We have outlined how to extend the {\itshape reparameterisation trick} to use common random numbers when performing the resampling step in a particle filter. This limits the discontinuities encountered when calculating gradients that are used in HMC and NUTS to propose new parameters within p-MCMC. We have applied these algorithms to three problems and show that using NUTS in this context can improve the mixing of the Markov chain compared to using MALA or HMC. We also compare different methods for resampling and show that using CRN resampling can produce more accurate estimates in shorter run time.

Although we have only included the methods for estimating the derivatives of Gaussian models, we note that our method could be applied to other models as long as the derivatives can be calculated. Considering a variety of different models would be a sensible direction for future work.

In this piece of work we have not included any analysis using Hessian information about the log-posterior within MCMC proposals, as was considered in \cite{Second-order}: This is due to the generic complexity of having to compute the second-order partial and full derivatives of the equations seen in sections \ref{sect:likelihoodandgradients} and \ref{sect:calcder}. 
An estimate of the state-dependent Hessian matrix could be made, using the gradients estimated in (\ref{eq:gradientlogposterior}), via the Gaussian Process construction provided in \cite{WILLS2021109503}. Hence, an avenue for future work is to include the resulting Hessian matrix in the proposal (\ref{MALA_eq}), along the lines of what was done in \cite{Second-order}, or as the mass matrix within NUTS. Doing so could provide additional performance gains over those reported herein.

An interesting direction for future work would involve a broader comparison of algorithms, that differ from p-MCMC, which can be applied to parameter estimation in SSMs. These include but are not limited to SMC$^2$ \cite{chopin2013smc2}, Nudging the Particle Filter \cite{akyildiz2020nudging} and the Nested Particle Filter \cite{crisan2018nested}.

The nested particle filter and SMC$^2$ have two layers of SMC method: one (an SMC sampler with $N_\theta$ particles) estimates the pdf over the static parameters, $\theta$, and the other (a particle filter with $N_x$ particles) considers the dynamic states. The difference between the two methods is that the nested particle filter runs in a purely recursive manner. A detailed comparison of the nested particle filter and SMC$^2$ can be seen in \cite{crisan2018nested} but the computational complexity of both methods is $O(N_\theta N_x T)$, just like the methods described in this paper (assuming we ran for $N_\theta$ MCMC steps). Future work would sensibly include a comparison with these alternative methods.

In \cite{akyildiz2020nudging} particles are {\it{nudged}} towards regions of the state space where the likelihood is deemed to be high. They present numerical improvements in terms of error rates in certain situations. They also apply this method to the particle Metropolis-Hastings algorithm \cite{andrieu_doucet_holenstein_2010} and outline how gradient nudging steps can be used within the framework of differentiable likelihoods and automatic differentiation libraries. This would be applicable to the methods proposed in this work with the potential to offer improvements in performance.

We also note that there is a trade-off between the theoretical concerns related to convergence and the empirical performance achieved. An avenue for future work would be to derive proofs related to the regularity properties of the estimated derivatives considered in this paper. 

\section*{Funding}

This work was supported by a Research Studentship jointly funded by the EPSRC and the ESRC Centre for Doctoral Training on Quantification and Management of Risk and Uncertainty in Complex Systems Environments [EP/L015927/1], the EPSRC through the Big Hypotheses grant [EP/R018537/1], the project  \emph{NewLEADS - New Directions in Learning Dynamical Systems} (contract number: 621-2016-06079), funded by the Swedish Research Council and by the \emph{Kjell och M{\"a}rta Beijer Foundation}, and a Research Studentship jointly funded by the EPSRC Centre for Doctoral Training in Distributed Algorithms [EP/S023445/1] and the UK Government.

\section*{Acknowledgments}

The authors would like to thank Lee Devlin and the anonymous reviewers for their insightful comments which improved the reading of this paper. 


%

\bibliographystyle{ieeetr}
\bibliography{bibliography.bib}

\appendices
\section{Partial versus total derivatives} \label{app:partial}
To try to avoid confusion, we use the partial derivative $\partial/\partial\theta$ to mean the derivative by only changing that function argument, and the total derivative $d/d\theta$ to mean also changing the other arguments depending on it, i.e.\ if $\theta$ is a scalar,
\begin{align}
	\frac{d}{d\theta}f(a(\theta), \theta) & \triangleq \mbox{lim}_{h\rightarrow 0}
		\frac{f(a(\theta + h), \theta + h) - f(a(\theta),\theta)}{h} \\
	\frac{\partial}{\partial \theta}f(a(\theta), \theta) & \triangleq \mbox{lim}_{h\rightarrow 0}
		\frac{f(a(\theta), \theta + h) - f(a(\theta),\theta)}{h}.
\end{align}

\section{Differentiating a Kalman Filter} \label{app:diffkalman}

We have a transition kernel and a measurement model as follows, where $\theta$ is a parameter vector:
\begin{align}
	p(\xp | x, \theta) & = \Normal(\xp; a(x, \theta), Q(x, \theta)) \\
	p(y | \xp, \theta) & = \Normal(y; h(\xp, \theta), R(\xp, \theta)).
\end{align}
Applying an Extended Kalman Filter gives a proposal for $\xp$ of the form
\begin{align}
	q(\xp | x, \theta, y) & = \Normal(\xp; \mu(x, \theta, y), C(x, \theta, y)).
\end{align}
We wish to calculate the derivatives
\begin{eqnarray} \nonumber
	\frac{\partial \mu}{\partial x}, \frac{\partial \mu}{\partial \theta}, \frac{\partial C}{\partial x}, \frac{\partial C}{\partial \theta}.
\end{eqnarray}
The standard Kalman filter equations are
\begin{align}
	S(x, \theta) & = H(a(x, \theta), \theta)Q(x, \theta)H(a(x, \theta), \theta)^T \\ \nonumber
	& \ \ \ + R(a(x, \theta), \theta) \\
	K(x, \theta) & = Q(x, \theta)H(a(x, \theta), \theta)^T S(x, \theta)^{-1} \\
	\mu(x, \theta, y) & = a(x, \theta) + K(x, \theta)(y - h(a(x, \theta), \theta)) \\
	C(x, \theta) & = Q(x, \theta) - K(x, \theta)H(a(x, \theta), \theta)Q(x, \theta)
\end{align}
where
\begin{align}
	H(a, \theta) & = \left(\frac{\partial h_i}{\partial a_j}(a, \theta)\right)_{ij}
\end{align}
is the Jacobian of the measurement function evaluated at the prior mean. We would like to differentiate these with respect to $x$ and $\theta$ but the
measurement model is defined in terms of the prior mean $a(x)$. Let
\begin{align}
	\ha(x, \theta) & = h(a(x, \theta), \theta) \\
	\Ha(x, \theta) & = H(a(x, \theta), \theta) \\
	\Ra(x, \theta) & = R(a(x, \theta), \theta).
\end{align}
Then
\begin{align}
	S(x, \theta) & = \Ha(x, \theta)Q(x, \theta)\Ha(x, \theta)^T + \Ra(x, \theta) \\
	K(x, \theta) & = Q(x, \theta)\Ha(x, \theta)^T S(x, \theta)^{-1} \\
	\mu(x, \theta, y) & = a(x, \theta) + K(x, \theta)(y - \ha(x, \theta))) \label{eqn:mu_x} \\
	C(x, \theta) & = Q(x, \theta) - K(x, \theta)\Ha(x, \theta)Q(x, \theta).	\label{eqn:C_x}
\end{align}
To compute the derivatives of these from the derivatives in $a$, applying the chain rule gives
\begin{align}
	\frac{\partial\ha}{\partial x}(x, \theta) & = \frac{\partial h}{\partial a}(a(x, \theta), \theta)\frac{\partial a}{\partial x}(x, \theta) \\
		& = H(a(x, \theta), \theta)\frac{\partial a}{\partial x}(x, \theta) \\
	\frac{\partial\ha}{\partial\theta}(x, \theta) & =  \frac{\partial h}{\partial a}(a(x, \theta), \theta)\frac{\partial a}{\partial \theta}(x, \theta) + 	
		\frac{\partial h}{\partial \theta}(a(x, \theta), \theta) \\
		& = H(a(x, \theta), \theta)\frac{\partial a}{\partial \theta}(x, \theta) + 	
			\frac{\partial h}{\partial \theta}(a(x, \theta), \theta) \\
	\frac{\partial\Ra}{\partial x}(x, \theta) & = \frac{\partial R}{\partial a}(a(x, \theta), \theta)\frac{\partial a}{\partial x}(x, \theta) \\
	\frac{\partial\Ra}{\partial\theta}(x, \theta) & =  \frac{\partial R}{\partial a}(a(x, \theta), \theta)\frac{\partial a}{\partial \theta}(x, \theta) + 	
		\frac{\partial R}{\partial \theta}(a(x, \theta), \theta) \\
	\frac{\partial\Ha}{\partial x}(x, \theta) & = \frac{\partial H}{\partial a}(a(x, \theta), \theta)\frac{\partial a}{\partial x}(x, \theta) \\ & =
		\frac{\partial^2 h}{\partial a^2}(a(x, \theta), \theta)\frac{\partial a}{\partial x}(x, \theta) \\
	\frac{\partial\Ha}{\partial\theta}(x, \theta) & = \frac{\partial H}{\partial a}(a(x, \theta), \theta)\frac{\partial a}{\partial \theta}(x, \theta) + 	
		\frac{\partial H}{\partial \theta}(a(x, \theta), \theta) \\ & =
	\frac{\partial^2 h}{\partial a^2}(a(x, \theta), \theta)\frac{\partial a}{\partial \theta}(x, \theta) + 	
		\frac{\partial^2h}{\partial a\partial \theta}(a(x, \theta), \theta)
\end{align}
Hence to evaluate the derivatives of (\ref{eqn:mu_x}) and (\ref{eqn:C_x}), we need
\begin{equation}
	a(x, \theta), \frac{\partial a}{\partial x}, \frac{\partial a}{\partial \theta}, Q(x, \theta), \frac{\partial Q}{\partial x},  \frac{\partial Q}{\partial\theta}
\end{equation}
from the transition model and
\begin{equation}
	h(a,\theta), \frac{\partial h}{\partial a}, \frac{\partial h}{\partial \theta}, \frac{\partial^2 h}{\partial a^2},
		\frac{\partial^2 h}{\partial a\partial\theta},  R(a(x),\theta), \frac{\partial R}{\partial a}, \frac{\partial R}{\partial\theta}
\end{equation}
from the measurement model. From this we apply the product rule and the inverse derivative in Appendix \ref{app:invder}.

\section{Derivatives of multivariate log normal}	\label{app:normalderiv}

If
\begin{align}
	\Normal(x; \mu, C) & = \frac{\exp\left(-\frac12(x - \mu)^TC^{-1}(x - \mu)\right)}{\sqrt{|2\pi C|}}
\end{align}
then
\begin{align}
	\frac{\partial}{\partial x}\logN  & = -C^{-1}(x - \mu) \label{eqn:dlogNdx} \\
	\frac{\partial}{\partial \mu}\logN & = C^{-1}(x - \mu) \label{eqn:dlogNdmu} \\
	\frac{\partial}{\partial C}\logN & = -\frac12\left(C^{-1} - C^{-1}(x - \mu)(x - \mu)^TC^{-1}\right). \label{eqn:dlogNC}
\end{align}

\section{Matrix derivatives}

\subsection{Derivative of a matrix inverse} \label{app:invder}

Suppose $U$ is an $N\times N$ invertible matrix with $N\times N$ derivative with respect to $\theta_r$ given by $dU/d\theta_r$. Then
\begin{align}
	\frac{\partial(U^{-1})}{\partial\theta_r} & =  -U^{-1}\left(\frac{\partial U}{\partial\theta_r}\right)U^{-1}.	\label{eqn:invder}
\end{align}
If $\theta$ is an $R$-dimensional vector, $d(U^{-1})/d\theta$ is an $N\times N\times R$ tensor with slice $r$ given by (\ref{eqn:invder}). 

\subsection{Derivative of a matrix square root}	\label{app:dsqrtm}

Suppose that $A$ is the matrix square root of $C$, i.e.
\begin{align}
	C & = AA.
\end{align}
Applying the product rule gives
\begin{align}
	\frac{\partial C}{\partial\theta} & = A\frac{\partial A}{\partial\theta} + A\frac{\partial A}{\partial\theta}
\end{align}
hence
\begin{align}
	\frac{\partial A}{\partial\theta} & = \frac{1}{2}A^{-1}\frac{\partial C}{\partial\theta}.
\end{align}

\end{document}